\documentclass[sigconf]{acmart}
\usepackage{subcaption}
\usepackage{multirow}

\AtBeginDocument{%
  \providecommand\BibTeX{{%
    \normalfont B\kern-0.5em{\scshape i\kern-0.25em b}\kern-0.8em\TeX}}}

\setcopyright{acmlicensed}

\copyrightyear{2024}
\acmYear{2024}
\setcopyright{acmlicensed}\acmConference[MM '24]{Proceedings of the 32nd ACM International Conference on Multimedia}{October 28-November 1, 2024}{Melbourne, VIC, Australia}
\acmBooktitle{Proceedings of the 32nd ACM International Conference on Multimedia (MM '24), October 28-November 1, 2024, Melbourne, VIC, Australia}
\acmDOI{10.1145/3664647.3681229}
\acmISBN{979-8-4007-0686-8/24/10}

\begin{document}

\title{Lite-Mind:\\ Towards Efficient and Robust Brain Representation Learning}
\renewcommand{\shorttitle}{Lite-Mind: Towards Efficient and Robust Brain Representation Learning}


\author{Zixuan Gong}
\affiliation{%
  \institution{Tongji University}
  \city{Shanghai}
  \country{China}}
\email{gongzx@tongji.edu.cn}

\author{Qi Zhang}
\affiliation{%
  \institution{Tongji University}
  \city{Shanghai}
  \country{China}}
\email{zhangqi\_cs@tongji.edu.cn}

\author{Guangyin Bao}
\affiliation{%
  \institution{Tongji University}
  \city{Shanghai}
  \country{China}}
\email{baogy@tongji.edu.cn}

\author{Lei Zhu}
\affiliation{%
  \institution{Tongji University}
  \city{Shanghai}
  \country{China}}
\email{leizhu0608@tongji.edu.cn}

\author{Ke Liu}
\affiliation{%
  \institution{Beijing Anding Hospital}
  \city{Beijing}
  \country{China}}
\email{liuke\_cist@mail.bnu.edu.cn}

\author{Liang Hu}
\affiliation{%
  \institution{Tongji University}
  \city{Shanghai}
  \country{China}}
\email{lianghu@tongji.edu.cn}

\author{Duoqian Miao}
\authornote{Corresponding author.}
\affiliation{%
  \institution{Tongji University}
  \city{Shanghai}
  \country{China}}
\email{dqmiao@tongji.edu.cn}

\author{Yu Zhang}
\affiliation{%
	\institution{Tongji University}
	\city{Shanghai}
	\country{China}}
\email{izy@tongji.edu.cn}

\renewcommand{\shortauthors}{Zixuan Gong, Qi Zhang, Guangyin Bao, Lei Zhu, Yu Zhang, Ke Liu, Liang Hu, Duoqian Miao*}

\begin{abstract}
  The limited data availability and the low signal-to-noise ratio of fMRI signals lead to the challenging task of fMRI-to-image retrieval. State-of-the-art MindEye remarkably improves fMRI-to-image retrieval performance by leveraging a large model, i.e., a 996M MLP Backbone per subject, to align fMRI embeddings to the final hidden layer of CLIP’s Vision Transformer (ViT). However, significant individual variations exist among subjects, even under identical experimental setups, mandating the training of large subject-specific models. The substantial parameters pose significant challenges in deploying fMRI decoding on practical devices. To this end, we propose \textbf{Lite-Mind}, a lightweight, efficient, and robust brain representation learning paradigm based on Discrete Fourier Transform (DFT), which efficiently aligns fMRI voxels to fine-grained information of CLIP. We elaborately design a DFT backbone with Spectrum Compression and Frequency Projector modules to learn informative and robust voxel embeddings. Our experiments demonstrate that \textbf{Lite-Mind} achieves an impressive 94.6\% fMRI-to-image retrieval accuracy on the NSD dataset for Subject 1, with 98.7\% fewer parameters than MindEye. \textbf{Lite-Mind} is also proven to be able to be migrated to smaller fMRI datasets and establishes a new state-of-the-art for zero-shot classification on the GOD dataset. 
\end{abstract}





\maketitle

\section{Introduction}
Brain decoding holds immense significance in elucidating intricate mental processes and serves as a pivotal technology for brain-computer interfaces~\cite{fMRIencoding,fMRI2005decoding,palazzo2020decoding}. Among diverse brain decoding tasks, the decoding of visual information stands out as a paramount yet challenging endeavor, allowing us to unravel the complex mechanisms involved in visual processing, object recognition, scene understanding, and even higher-order cognitive functions. In the pursuit of decoding natural visual information, functional magnetic resonance imaging (fMRI) is widely employed as a non-invasive modality to decipher the perceptual and semantic information intricately encoded within the cerebral cortex~\cite{fMRIbook}. It draws considerable attention to fMRI-to-image retrieval/reconstruction tasks. This paper focuses primarily on tasks related to fMRI-to-image retrieval.

\begin{figure}[t]
  \centering
   \includegraphics[width=1.0\linewidth]{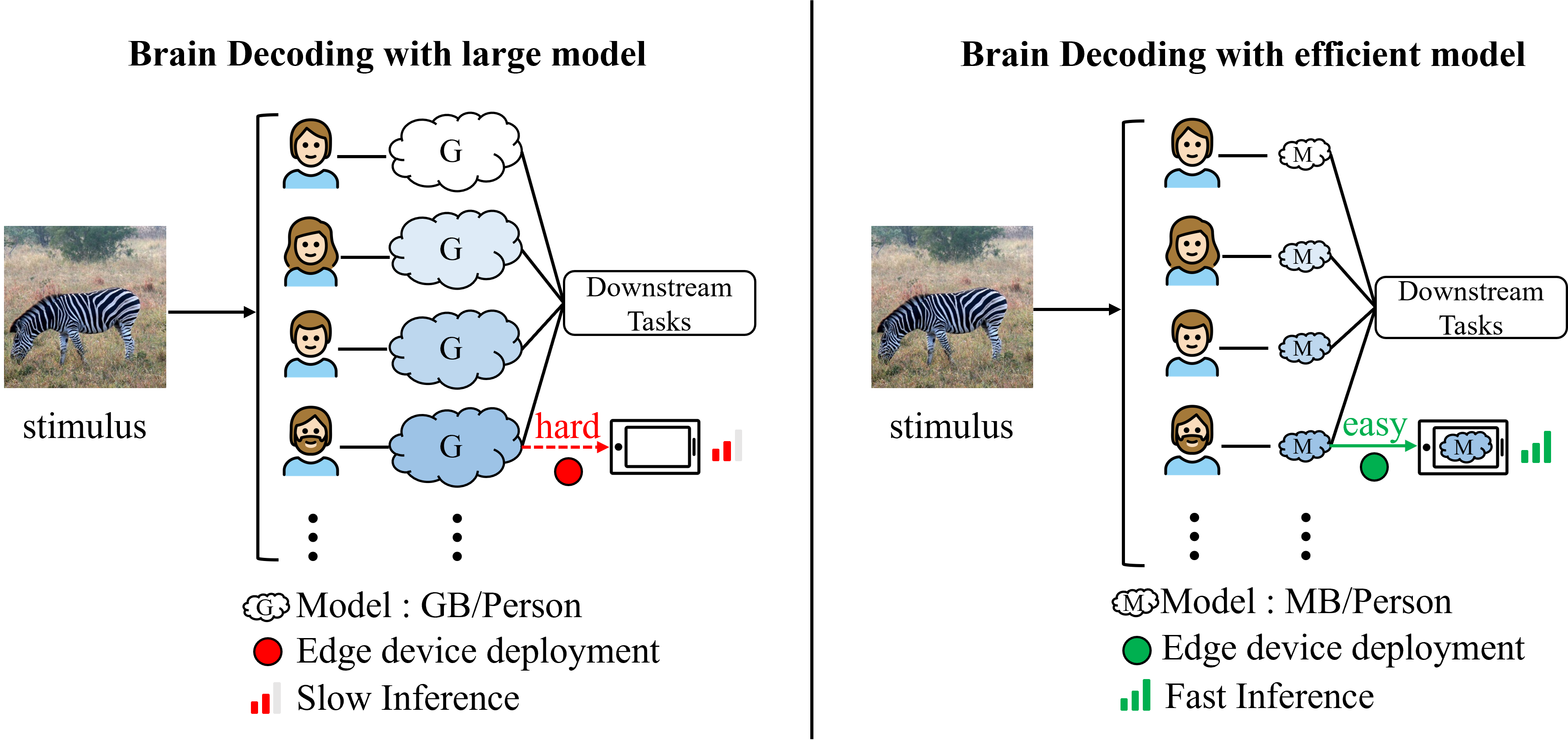}
   \vspace{-5mm}
   \caption{The significance of lightweight brain decoding model. Subject-specific models are visually represented by clouds of distinct colors, and the cloud size correlates with model parameters.} 
   \label{fig:intro}
   \vspace{-6mm}
\end{figure}

The success of fMRI-to-image retrieval heavily relies on aligning fMRI signals with the representation space of images, following a series of cross-modal alignment efforts~\cite{ramesh2022dalle2,wu2024hydiscgan,clip}. However, fMRI data often suffers from spatial redundancy, noise, and sample sparsity, leading to poor representation of fMRI signals and potential overfitting to noise distribution for deep models. To address these challenges, previous studies have found ridge regression and shallow linear models to be effective solutions for fMRI-based models~\cite{r3,r5,emmanuele2021decoding,nsd,horikawa2017god,corsscloze}. Recently, it has been demonstrated the effectiveness of leveraging pretrained CLIP models~\cite{clip} as a powerful bridge between fMRI voxels and images, since CLIP's image embeddings capture fine-grained and semantic information of pictures~\cite{liu2023brainclip,mai2023unibrain,mindreader2022,mindseye2023,zhang2024mlip}. Initially, the CLS embeddings of the CLIP image encoder was used to align voxels in Mind-Reader~\cite{mindreader2022}. 
BrainClip~\cite{liu2023brainclip} and UniBrain~\cite{mai2023unibrain} introduce the text annotations of images to assist in aligning fMRI voxels with image representations. Brain-diffuser~\cite{brain-diffuser} aligns voxel representations with the last hidden layer of CLIP using large-scale linear models ($257\times14$M) for the first time.

Among the CLIP-related models, MindEye~\cite{mindseye2023} stands out by employing a large-scale (996M) MLP Backbone and Projector to map fMRI voxels and align their representations with the final hidden layer of CLIP using contrastive learning. MindEye achieves remarkable breakthroughs, surpassing $90\%$ accuracy for both image and brain retrieval, outperforming previous state-of-the-art accuracies that were below $50\%$. However, the practical application of large-scale models in brain science research faces limitations. Each subject's fMRI data can exhibit significant variations, even within identical experimental setups, making it currently infeasible to establish a universal encoding model that performs well for all individuals. Consequently, training individual encoding models for each subject becomes necessary, as depicted in Figure \ref{fig:intro}. Apparently, it is impractical to expect hospitals to have access to computing resources with sufficient performance to train a unique large-scale model (e.g., MindEye) for each individual. Therefore, the ideal scenario involves equipping individuals with lightweight brain-computer interface edge devices, enabling image representations to be efficiently retrieved by a lightweight model. To this end, there is an urgent need to develop a brain network that is lightweight and efficient, as illustrated in the right segment of Figure \ref{fig:intro}, for practical implementation and deployment of brain-computer interfaces in real-world scenarios.

MindEye strongly benefits from an MLP backbone with a high parameter count, which allows for straightforward and powerful compression and preservation. However, when dealing with fMRI data, which typically has a low signal-to-noise ratio, an MLP focusing on voxel value-wise mappings may inadvertently suppress informative voxel signal values while being inefficient at reducing noisy values. This can lead to model parameter redundancy and decelerated convergence speed. Recently, the Fourier Transform has gained attention in deep learning, demonstrating lightweight characteristics and improved efficiency in learning frequency domain patterns, due to its global perspective and energy concentration properties~\cite{wang2023fftmlp}. Additionally, the Fourier Transform naturally possesses the advantage of processing signals: it is more effective and efficient to filter out noise and maintain informative voxel signals in the frequency domain. It presents a promising alternative to large-scale MLPs for encoding fMRI signals.

In order to mitigate the noise effect of fMRI and obtain more robust latent representations, we designed global filter-based Filter Blocks for fMRI denoising and fMRI spectrum compression. In addition to the global view property of frequency domain filtering, which can block out the noise and thus obtain better robust representations, the frequency domain computation also brings a great efficiency improvement compared to the large MLP structure. Since the features of the compressed fMRI we obtain are more concentrated in the frequency domain, we also design a Frequency Projector based on MLP in the frequency domain. Combining all our computational paradigms in frequency domain, we propose \textbf{Lite-Mind}, a lightweight, efficient, and robust brain representation learning paradigm. We redesign MindEye with our elaborate Discrete Fourier Transform (DFT) Backbone, avoiding the large MLP Backbone used in MindEye. Extensive experiments show that \textbf{Lite-Mind} achieves 94.6\% retrieval accuracy for Subject 1 on the NSD dataset, with 98.7\% fewer parameters than MindEye. \textbf{Lite-Mind} also proves its adaptability to smaller brain datasets and establishes a new state-of-the-art zero-shot classification on the GOD dataset. Our contributions are summarized as follows:

\begin{itemize}
\item We demonstrated that the Fourier Transform has the advantage of being lightweight and efficient for fMRI representation learning in the frequency domain.
\item We proposed a novel  \textbf{DFT Backbone} with elaborate Spectrum Compression and Frequency Projector modules, which is theoretically and empirically verified lightweight, efficient, and robust for brain representation.
\item We migrated \textbf{Lite-Mind} to various downstream tasks, and demonstrated its robustness. It achieves state-of-the-art zero-shot classification performance on the GOD dataset.
\end{itemize}

\section{Related Work}
\label{sec:formatting}
\subsection{Brain Visual Decoding}
The study of visual decoding using fMRI in the human brain has been a long-standing endeavor. In the most important fMRI-image reconstruction and partial fMRI-image retrieval tasks, the foundation is how to align fMRI signals to the image representation space or intermediate space. Due to the low signal-to-noise ratio of fMRI, the initial research emphasized using models based on linear regression to extract fMRI signals and images into the intermediate space for functional decoding of the human brain or reconstruction of faces and natural scenes~\cite{takagi2023high}. Further research utilizes pre-trained VGG to enrich hierarchical image features~\cite{horikawa2017god,shen2019deep,pami}. With the development of cross-modal tasks, the image-text pre-training model CLIP has been introduced into fMRI-image research. Mind-reader~\cite{mindreader2022} is the first to adopt a contrastive learning approach, using the \textit{nsdgenal} ROI region of NSD data and a shallow Resnet-like model to align fMRI with the CLS embeddings output by the CLIP image encoder. Mind-Reader did not achieve good performance in fMRI-image retrieval when using InfoNCE loss. BrainClip~\cite{liu2023brainclip} uses a VAE model for retrieval tasks while still using the CLS embeddings. Brain diffuser~\cite{brain-diffuser} introduced a fine-grained representation of the image of the last hidden layer in CLIP. However they used 257 different linear regression models to align each layer of the 257 hidden layers, resulting in the model being too cumbersome. Mind-Vis~\cite{chen2023seeing} conducted self-supervised mask pre-training on fMRI signals for another dataset, BOLD5000, and demonstrated that voxel patches can be used to process fMRI voxel values. MindEye~\cite{mindseye2023} uses a large MLP backbone for contrastive learning, aligning the last hidden layer $257\times768$ of CLIP, and using diffusion prior in DALLE·2~\cite{ramesh2022dalle2} for the first time to narrow the disjointed vectors. At the same time, it creatively proposed a mixed contrast loss MixCo as a data augmentation method to make the large model converge. The recent trend in brain fMRI decoding research tends to use larger models to achieve better performance on downstream tasks~\cite{xia2024dream,kneeland2023boi}, ignoring the privacy and deployment issues discussed above. However, to the best of our knowledge, no previous research has focused on lightweight and efficient brain networks.

\begin{figure*}[!t]
  \centering
  \includegraphics[width=1.0\linewidth]{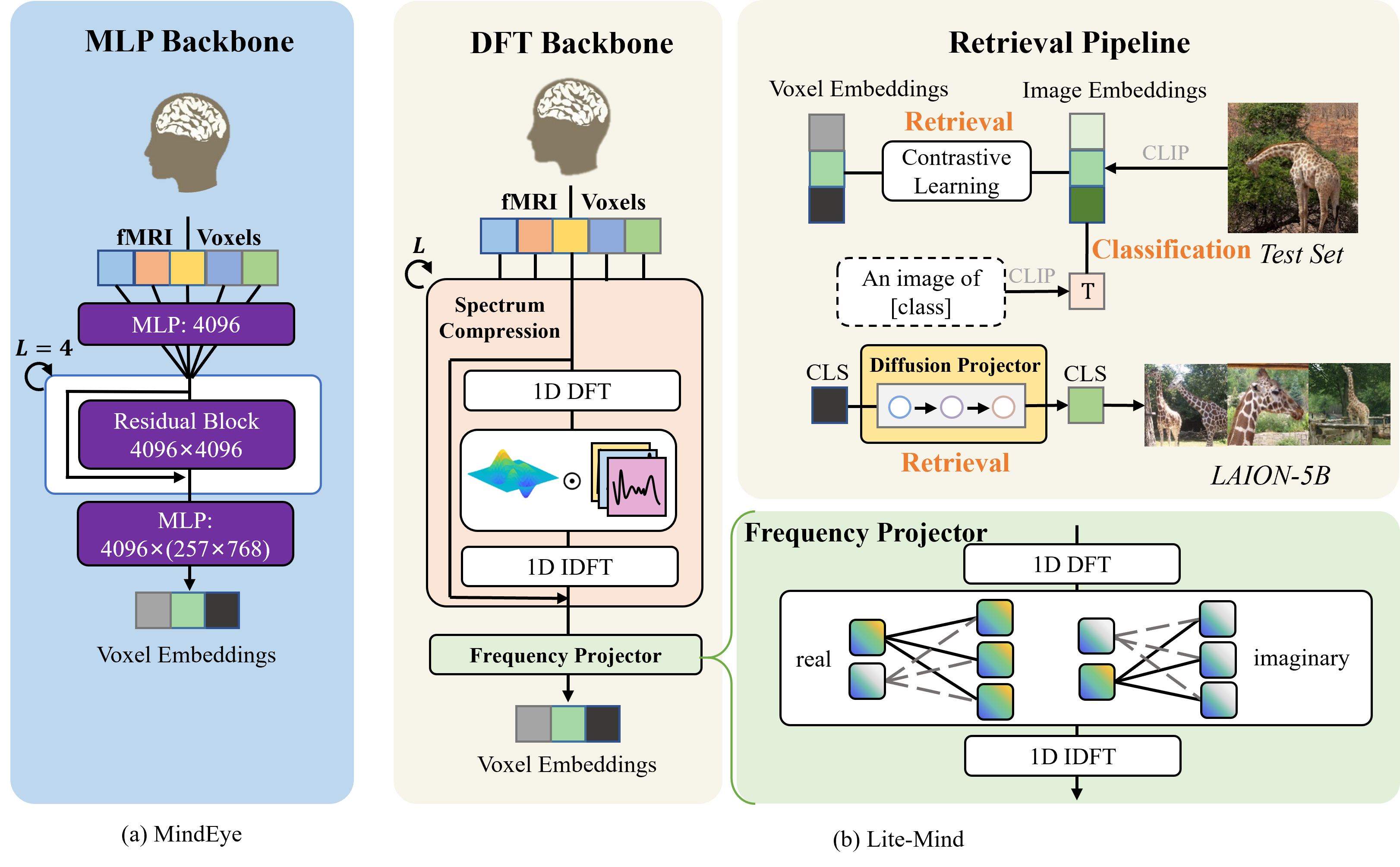}
  \vspace{-8mm}
   \caption{Overview of our Lite-Mind. Figures (a) and (b) show the architecture of the MLP Backbone of MindEye and the DFT backbone and Retrieval Pipeline, respectively. fMRI voxels are inputted into DFT Backbone to obtain voxel embeddings.}
   \label{fig:model}
   \vspace{-4mm}
\end{figure*}

\subsection{Fourier Transform in Deep Learning}
Fourier Transform plays a vital role in the area of digital signal processing. It has been introduced to deep learning for enhanced learning performance~\cite{yi2023surveydeeplearningbased, LaoZSCYHM24, yang2020fda, zhang2024mlip}, and has demonstrated excellent performance in many fields, such as computer vision, natural language processing, and time series analysis. In the field of computer vision, GFNet~\cite{gfnet} utilizes Fast Fourier Transform (FFT) to convert images to the frequency domain and exchange global information between learnable filters. As a continuous global convolution independent of input resolution, Guibas et al.~\cite{guibas2021adaptive} design the Adaptive Fourier Neural Operator (AFNO) frame token mixing. Xu et al.~\cite{xu2020learning} devise a learning-based frequency selection method to identify trivial frequency components and improve image classification accuracy. As for natural language processing, Lee-Thorp et al.~\cite{lee2021fnet} use the Fourier Transform as a text token mixing mechanism for text classification. In research of time series forecast, Fourier Transform has deeper applications ~\cite{YiZFHHWACN23,lange2021fourier,kocc2022fractional}. To increase the accuracy of multivariate time-series forecasting, Cao et al.~\cite{cao2020spectral} propose a spectral temporal graph neural network (StemGNN), which mines the correlations and time dependencies between sequences in the frequency domain. Yang et al.~\cite{yang2022unsupervised} propose bilinear temporal spectral fusion (BTSF), which updates the feature representation in a fused manner by explicitly encoding time-frequency pairs and using two aggregation modules: spectrum-to-time and time-to-spectrum. Yi et al.~\cite{wang2023fftmlp} prove that frequency-domain MLP is more efficient than time-domain MLP. Processing fMRI frequency domain information is more helpful for understanding the mechanism of collaborative operation within the human brain. However, there is currently no research on processing fMRI signals in the frequency domain for representation learning.

\section{Lite-Mind}
\subsection{Overview}
As illustrated in Figure \ref{fig:model}, Lite-Mind comprises two main components: 1) \textbf{DFT Backbone} consists of \textit{fMRI Spectrum Compression} and \textit{Frequency Projector}, mapping flattened voxels to an intermediate space. 2) \textbf{Retrieval pipeline} contains \textit{Diffusion Projector}, \textit{Contrastive Learning}, and \textit{Downstream tasks} with the voxel embeddings from the DFT backbone.
\subsection{DFT Backbone}
Sampling an fMRI-image pair ($x$,$y$) from dataset $D$, $x$ denotes the fMRI data and $y$ denotes the paired image. 

\subsubsection{fMRI Spectrum Compression}

\textbf{Patchify and Embedding} The fMRI data $x$ after ROI preprocessing and flattening is a one-dimensional long vector composed of voxels. We conduct patchify firstly because it has been verified as simple and effective for spatial/temporal representation of long sequences or high-dimensional vectors. At first, we divide $x$ into $n$ non-overlapping patches $x = [x_1, x_2,...,x_n]$ using zero-padding, and adopt convolutions with positional encoding to obtain multiple tokens $t = [t_1, t_2,...,t_n]$.

\textbf{Filter Block.} The spectrum of voxel tokens is obtained by 1D DFT as follows:
\begin{equation}
    X[k] = F(t) = \sum_{i=1}^{n}t_{i} e^{-ki(2\pi/n)j}
\end{equation}
Where $X\in\mathbb{C}^{n\times d}$ is a complex tensor, $X[k]$ is the spectrum of $t$ at the frequency $2\pi k/n$, $F(\cdot)$ denotes the 1D DFT along the voxel token dimension. $i$ denotes the $i$-th token, and $j$ denotes unit imaginary.



Voxel spatial features are effectively consolidated within each element of the frequency spectrum of voxel tokens, enabling the extraction of informative features from voxels through the point-wise product in the frequency domain. Accordingly, we introduce learnable filters $\textbf{K} = [\textbf{k}_1,\textbf{k}_2,...,\textbf{k}_M]$ to filter and compress the spectrum as follows, where $\mathbf{K}$ denotes the filter library and $M$ is the number of filters in the library.
\begin{equation}
    \hat{X} = \sum_{m=1}^{M}\frac{1}{n}{|X|}^2 \odot \textbf{k}_{m} cos(\frac{(2m-1)\pi}{2M})
\end{equation}
where $\odot$ is the element-wise multiplication, $|X|^2$ is the power spectrum of $X$.
The $|X|^2$ operation smooths the spectrum, highlighting the main components of the fMRI spectrum from a global perspective. $cos((2m-1)\pi/2M)$ adopts the filtering form of Discrete Cosine Transform (DCT), which compacts better energy and can aggregate the significant information in fMRI voxels. Its combination with the channel filter library $\textbf{K}$ allows for efficient feature compression and noise denoising for fMRI from the frequency domain.


Finally, we employ Inverse Discrete Fourier Transform (IDFT) $F^{-1}(\cdot)$ to convert the spectrum back into the spatial domain:
\begin{equation}
    \hat{t} \leftarrow F^{-1}(\hat{X})
\end{equation}

\subsubsection{Frequency Projector}
The fMRI spectrum after filtering has a more concentrated feature in the frequency domain compared to the spatial domain, which is caused by the energy concentration effect of the frequency domain calculation. We therefore designed a Frequency Projector in the frequency domain to align the image tokens, which has the added benefit of maintaining all computations of our DFT Backbone in the frequency domain, where lightness and efficiency are guaranteed. To align voxel tokens to image tokens, we mapped the filtered and compressed fMRI tokens by an MLP-like projector in the frequency domain(FreMLP). Similarly, we follow Equation (2) to convert $\hat{t}$ to the frequency domain.

\textbf{FreMLP.} Following Equation (2), the complex number output of $\hat{t}$ after DFT is $\hat{X}\in\mathbb{C}^{n\times d}$. Given a complex number weight matrix $\mathcal{W}\in\mathbb{C}^{n\times n'}$, and a complex number bias $\mathcal{B}\in\mathbb{C}^{n'}$, then the FreMLP can be formulated as:
\begin{equation}
    X' = {\sigma(\hat{X}^T\mathcal{W}+\mathcal{B})}^T
\end{equation}
Where $X'\in\mathbb{C}^{n'\times d}$ is the final output, and $\sigma$ is the activation function. As both $\hat{X}$ and $\mathcal{W}$ are complex numbers, according to the rule of multiplication of complex numbers(see Appendix D.1 for details), we further extend the Equation (5) to:
\begin{equation}
    \begin{split}
    X' &= ({\sigma(Re(\hat{X}^T)\mathcal{W}_r-Im(\hat{X}^T)\mathcal{W}_i+\mathcal{B}_r)}\\
    &+j\sigma(Re(\hat{X}^T)\mathcal{W}_i+Im(\hat{X}^T)\mathcal{W}_r+\mathcal{B}_i))^T
    \end{split}
\end{equation}

Where $Re(\cdot)$ denotes the real part of $\hat{X}^T$, $Im(\cdot)$ denotes the imaginary part, $\mathcal{W}=\mathcal{W}_r+j\mathcal{W}_i$ and $\mathcal{B}=\mathcal{B}_r+j\mathcal{B}_i$. Due to the more significant features of filtered and compressed tokens in the frequency domain, we replace the final FC layer of MindEye's MLP Backbone with FreMLP. For more theoretical proof and implementation of FreMLP, please refer to Appendix D.1. Finally, we follow IDFT $t' \leftarrow F^{-1}(X')$ to get final tokens $t'$, and use voxel embedding $f$ to represent them.

\begin{figure*}[!t]
  \centering
  \vspace{-4mm}
   \includegraphics[width=1.0\linewidth]{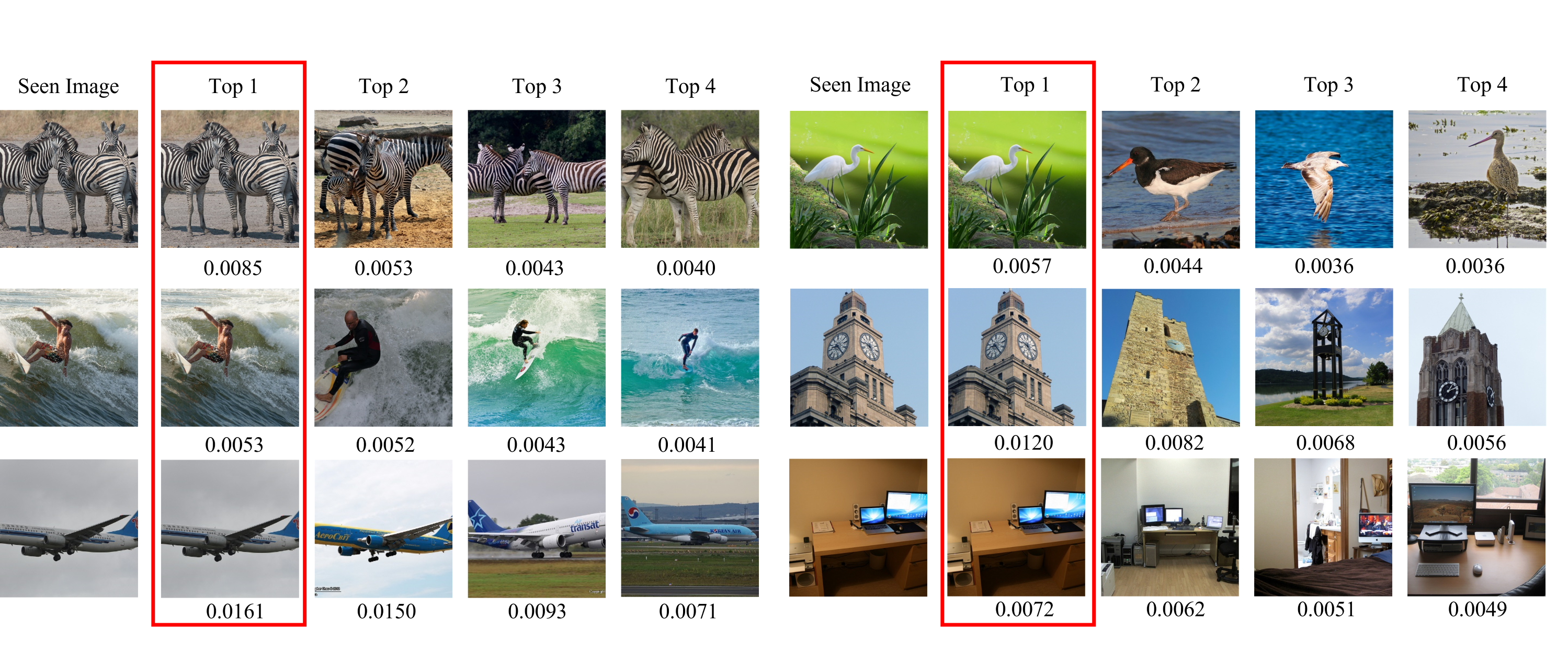}
\vspace{-7mm}
   \caption{Partial retrieval results of Lite-Mind on all 982 test images for Subject 1. With 12.5M DFT Backbone, Lite-Mind can still find the exact Top-1 image pair from the test set of 982 images with 89.3\% accuracy and 95.5\% accuracy of image-fMRI retrieval(random chance = 0.1\%) and can distinguish among confusable candidates. The number below each image represents the similarity score. See more cases including failure retrieval in Appendix C.4.}
   \vspace{-3mm}
   \label{fig:retrieval result}
\end{figure*}

\subsection{Retrieval Pipeline}

\subsubsection{Optimization Objective}
The optimization objective for the retrieval task is as follows:
\begin{equation}
\omega^{*} = \underset{\omega}{argmax} \sum_{(x,y)\in D}SIM(DFT(x;\omega),CLIP(y))
\end{equation}
where $\omega$ is the weight of our DFT backbone, $SIM(\cdot)$ denotes the cosine similarity, and $CLIP(\cdot)$ denotes using CLIP model to extract image embedding.

\subsubsection{Diffusion Projector}
When searching on a large-scale dataset like LAION-5B, image-to-image performs better than fMRI-to-image and can find more similar images. Therefore, we used a diffusion projector to translate the disjointed voxel CLS embeddings of the DFT Backbone output, using a DALLE · 2\cite{ramesh2022dalle2} similar diffusion model to obtain the image CLS embeddings and retrieve it on LAION-5B: $\mathcal{V'} = Diffusion(f)$. The Diffusion Projector consists of a transformer for image CLS embedding forecasting.

\subsubsection{Contrastive Learning}
Due to the inherent differences between different modalities, we use CLIP's contrastive loss to train the DFT backbone. Sampling a batch $B$ from voxel-image pairs, the contrastive loss is defined below:
\begin{equation}
  L_{contr}=-\frac{1}{|B|}
  \sum_{s=1}^{|B|} \log \frac
  {\exp(f_s^{\top}\cdot\mathcal{V}_s/{\tau})}
  {\sum_{i=1}^{|B|} \exp({f_s^{\top}\cdot\mathcal{V}_i}/{\tau})},
\end{equation}
where $f_s$ is the $s$-th voxel embeddings, $\mathcal{V}_s$ is its corresponding image embeddings, $\tau$ is a temperature factor.

To perform image-to-image retrieval on LAION-5B, we use MSE loss to constrain the generation of approximate image embeddings:

\begin{equation}
  L_{mse}=\frac{1}{|B|} \sum_{s=1}^{|B|} {||\mathcal{V}_s - \mathcal{V'}_s||}^2_2
\end{equation}

Accordingly, the training loss is defined below:
\begin{equation}
    L=L_{contr} + \alpha L_{mse}
\end{equation}
In general, when $\alpha = 0$, Lite-Mind is trained to output voxel embeddings for retrieval on the NSD test set. However, for LAION-5B retrieval, $\alpha$ is non-zero to map the voxel CLS embeddings outputted by Lite-Mind to image CLS embeddings, enabling online retrieval.


\subsubsection{Downstream tasks}
The retrieval process on various downstream tasks is shown in the upper right of Figure \ref{fig:model}.

\noindent\textbf{Test set retrieval.} The cosine similarities between voxel embeddings from DFT Backbone and image embeddings from CLIP ViT are directly calculated. 

\noindent\textbf{LAION-5B retrieval.} Voxel CLS embeddings from the DFT backbone are translated to image CLS embeddings through the Diffusion Projector for online LAION-5B retrieval.

\noindent\textbf{Zero-shot classification.} Image retrieval is performed on novel classes on the test set, and the similarities between the retrieved images and simple CLIP class text prompts are calculated for classification tasks.

\section{Experiments}
\subsection{Dataset}
Natural Scenes Dataset (NSD) is an extensive 7T fMRI dataset gathered from 8 subjects viewing images from MSCOCO-2017 dataset~\cite{nsd}, which contains images of complex natural scenes. Participants viewed three repetitions of 10,000 images with a 7-Tesla fMRI scanner over 30–40 sessions. More details can be found on the NSD official website\href{https://naturalscenesdataset.org}{}\footnote{\url{https://naturalscenesdataset.org}}. Our experiments focused on Subj01, Subj02, Subj05, and Subj07, who finished all viewing trials. Each subject's training set comprises 8859 image stimuli and 24980 fMRI trials (with the possibility of 1-3 repetitions per image). The test set contains 982 image stimuli and 2770 fMRI trials. Responses for images with multiple fMRI trials are averaged across these trials. By applying the \textit{nsdgeneral} ROI mask with a 1.8 mm resolution, we obtained ROIs for the four subjects, comprising 15724, 14278, 13039, and 12682 voxels respectively. These regions span visual areas ranging from the early visual cortex to higher visual areas. Our experimental setup is consistent with the NSD image reconstruction and retrieval articles~\cite{takagi2023high,brain-diffuser,mai2023unibrain,mindseye2023,mindreader2022}.

Generic Object Decoding (GOD) Dataset\href{http://brainliner.jp/data/brainliner/Generic_Object_Decoding}{}\footnote{\url{http://brainliner.jp/data/brainliner/Generic_Object_Decoding}} was created by Horikawa and Kamitani~\cite{horikawa2017god}, consisting of fMRI recordings of five healthy subjects who were presented with images from ImageNet~\cite{deng2009imagenet}. The GOD Dataset includes 1250 distinct images selected from 200 ImageNet categories. Among these, 1200 training images are drawn from 150 categories, and 50 test images are from the remaining 50 categories. The training and test image stimuli were presented to the subjects once and 35 times respectively, resulting in 1200 and 1750 fMRI instances. We used preprocessed ROIs, encompassing voxels from the early visual cortex to higher visual areas. For each test image, the fMRI responses from different trials were averaged.

\subsection{Implementation details}
To fairly compare different backbones, we did not use any data augmentation methods, such as voxel mixing loss MixCo and image slicing enhancement. Due to MindEye only disclosing the MLP Backbone performance of Subject 1, only the performance of Subject 1 is provided in Table \ref{tab2: main result}. The experimental results of more subjects are shown in Appendix B.1. All of our experiments were conducted on a single Tesla V100 32GB GPU. More experimental details and hyperparameter settings can be found in Appendix A.
\captionsetup[subtable]{font=small}
\begin{table*}[h]
  \centering
  \begin{subtable}{1.0\linewidth}
  \centering
  \begin{tabular}{@{}cccccccc@{}}
    \toprule
    \multirow{2}{*}{Method} & \multirow{2}{*}{Model}& \multirow{2}{*}{Parameters} & \multicolumn{2}{c}{Retrieval}\\
    \cline{4-5}
    & & & Image$\uparrow$&Brain$\uparrow$\\
    \midrule
     Lin et al.~\cite{mindreader2022}&deep models& 2.34M&11.0\% & 49.0\%\\
     Ozcelik...~\cite{brain-diffuser}&257 separate linear regression models&3B& 29.9\% & 21.4\%\\
     MindEye~\cite{mindseye2023}&MLP Backbone&940M& 89.6\% & 82.2\%\\
     MindEye~\cite{mindseye2023}&MLP Backbone+Projector&996M& 88.8\% & 84.9\%\\
     MindEye~\cite{mindseye2023}&MLP Backbone+Prior&1.2B&93.4\%&90.1\%\\
     Lite-Mind(ours)&DFT Backbone&\textbf{12.5M}&\textbf{94.6}\% & \textbf{97.1}\%\\
    \bottomrule
  \end{tabular}
  \caption{Quantitative comparison of Lite-Mind retrieval performance against other models for Subject 1. Image retrieval refers to the hit rate of correct retrieval from 300 candidates, given the associated brain sample (chance=0.3\%); vice-versa for brain retrieval. Lite-Mind only uses a parameter quantity of 12.5M, achieving extremely high retrieval performance without using any model to close vectors of different modalities (see Appendix B.1 for remaining subject models).}
  \label{tab2: main result}
  \end{subtable}
  \begin{subtable}{1.0\linewidth}
  \centering
  \begin{tabular}{@{}ccccccccc@{}}
    \toprule
    \multirow{2}{*}{Method} & \multicolumn{4}{c}{Low-Level}& \multicolumn{4}{c}{High-Level}\\
    \cmidrule(lr){2-5} \cmidrule(lr){6-9}
     &PixCorr$\uparrow$&SSIM$\uparrow$&Alex(2)$\uparrow$&Alex(5)$\uparrow$&Incep$\uparrow$&CLIP$\uparrow$&Eff$\downarrow$&SwAV$\downarrow$\\

    \midrule
     Lin et al.~\cite{mindreader2022}&-&-&-&-&78.2\%&-&-&-\\
     Takagi...~\cite{takagi2023high}&-&-&83.0\%&83.0\%&76.0\%&77.0\%&-&-\\
     Gu et al.~\cite{gu2022decoding}&.150&.325&-&-&-&-&.862&.465\\
     Ozcelik...~\cite{brain-diffuser}&.254&\textbf{.356}&94.2\%&96.2\%&87.2\%&91.5\%&.775&.423\\
     BrainCLIP~\cite{liu2023brainclip}&-&-&-&-&86.7\%&\textbf{94.8}\%&-&-\\
     MindEye~\cite{mindseye2023}&\textbf{.309}&.323&\textbf{94.7}\%&\textbf{97.8}\%&\textbf{93.8}\%&94.1&\textbf{.645}&\textbf{.367}\\
     \cmidrule(lr){1-9}
     MindEye-125M(LAION-5B)&.130&.308&84.0\%&92.6\%&86.9\%&86.1\%&.778&.477 \\
     Lite-Mind-8M(LAION-5B)&.125&\textbf{.331}&78.7\%&89.4\%&\textbf{87.9}\%&\textbf{88.7}\%&\textbf{.724}&\textbf{.446}\\
     \bottomrule
  \end{tabular}
  \caption{LAION-5B retrieval performance. The upper part is based on various fMRI-image reconstruction methods using generative neural networks, while the lower part represents the retrieval performance of the large-scale dataset LAION-5B as an alternative to fMRI-image reconstruction. All results are averaged across four subjects, and Lite-Mind achieved higher performance with a lighter backbone, even exceeding the accuracy of many reconstruction methods.}
  \label{tab2: LAION-5B}
  \end{subtable}
  \vspace{-6mm}
  \caption{Main results of Lite-Mind retrieval and LAION-5B retrieval on the NSD dataset.}
  \vspace{-7mm}
\end{table*}

\section{Results}
\subsection{fMRI/image retrieval}
Image retrieval refers to retrieving the image embeddings with the highest cosine similarity based on voxel embeddings on the test set. If a paired image embedding is retrieved, the retrieval is considered correct. fMRI retrieval is the opposite process mentioned above. Note that there are many semantically and visually similar images on the NSD test set, considered to be similar in CLIP space. Whether the model can correctly retrieve as MindEye tests the fine-grained alignment ability to the image. For test retrieval, we adhered to the identical methodology as Lin et al.~\cite{mindreader2022} and MindEye~\cite{mindseye2023} to compute the retrieval metrics presented in Table \ref{tab2: main result}. For each test sample, we randomly selected 299 images from the remaining 981 images in the test set and calculated the cosine similarity between the voxel embeddings and 300 images. The retrieval accuracy refers to the proportion of successful retrieval of corresponding images in the 982 voxel embeddings of the test set. We adjusted the random number seed of 30 randomly selected images to average the accuracy of all samples. The experimental results are shown in Table \ref{tab2: main result}. We also conducted retrieval experiments on the remaining three subjects to demonstrate the universality of DFT Backbone. The detailed results and discussions are in Appendix B.2.

\begin{figure*}[!t]
  \centering
   \vspace{-1mm}
\includegraphics[width=1.0\linewidth]{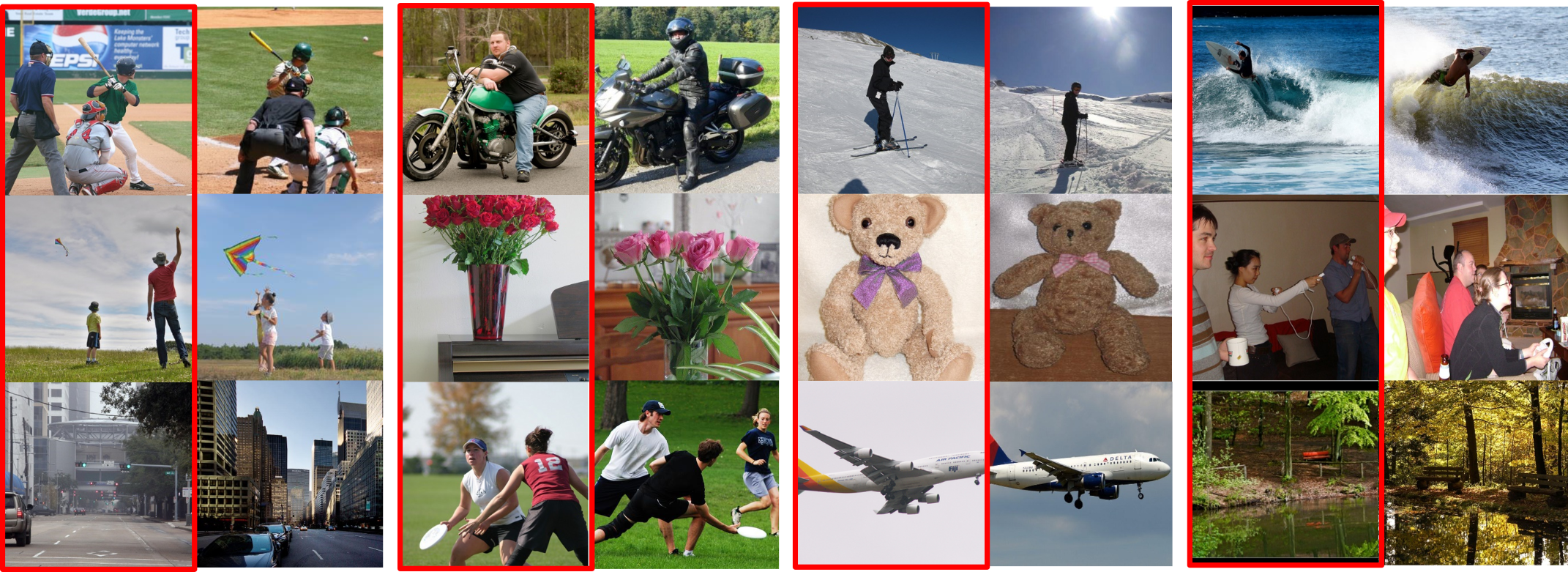}
   \vspace{-6mm}
   \caption{Retrieval results of Lite-Mind from LAION-5B for Subject 1. The left column marked by a red box in every two columns represents the original image seen by the Subject, and the right column represents the image retrieved from LAION-5B.}
   \vspace{-1mm}
   \label{fig:LAION-5B}
\end{figure*}

\begin{table*}[!t]
  \centering
  \vspace{-1mm}
  \scalebox{0.9}{
  \begin{tabular}{@{}ccccccccccccc@{}}
    \toprule
    \multirow{2}{*}{Methods} & \multirow{2}{*}{Modality} & \multirow{2}{*}{Prompt} & \multicolumn{2}{c}{Subject1} & \multicolumn{2}{c}{Subject2} & \multicolumn{2}{c}{Subject3} & \multicolumn{2}{c}{Subject4} & \multicolumn{2}{c}{Subject5}\\
    \cmidrule(lr){4-5} \cmidrule(lr){6-7} \cmidrule(lr){8-9} \cmidrule(lr){10-11} \cmidrule(lr){12-13} 
    & & &top-1&top5&top-1&top5&top-1&top5&top-1&top5&top-1&top5\\
    \midrule
    CADA-VAE~\cite{cada-vae}&{V\&T}&-&6.31&35.70&6.45&40.12&17.74&54.34&12.17&36.64&7.45&35.04\\
    MVAE~\cite{mvae}&{V\&T}&-&5.77&31.51&5.40&38.46&17.11&52.46&14.02&40.90&7.89&34.63\\
    MMVAE~\cite{mmvae}&{V\&T}&-&6.63&38.74&6.60&41.03&22.11&56.28&14.54&42.45&8.53&38.14\\
    MoPoE-VAE~\cite{mopoevae}&{V\&T}&-&8.54&44.05&8.34&48.11&22.68&61.82&14.57&58.51&10.45&46.40\\
    BraVL~\cite{pami}&{V\&T}&-&9.11&46.80&8.91&48.86&24.00&62.06&15.08&60.00&12.86&47.94\\
    BrainClip-VAE~\cite{liu2023brainclip}&{V\&T}&Text&8.00&42.00&24.00&52.00&20.00&58.00&20.00&58.00&20.00&46.00\\
    BrainClip-VAE~\cite{liu2023brainclip}&{V\&T}&CoOp&14.00&54.00&20.00&60.00&21.33&64.67&16.67&66.67&18.00&52.00\\
    MindEye-211M~\cite{mindseye2023}&{V}&Text&6.00&40.00&20.00&60.00&28.00&70.00&14.00&54.00&10.00&64.00\\
    Lite-Mind-15.5M(ours)&V&Text&\textbf{26.00}&\textbf{74.00}&\textbf{28.00}&\textbf{70.00}&\textbf{30.00}&\textbf{80.00}&\textbf{34.00}&\textbf{76.00}&\textbf{24.00}&\textbf{66.00}\\
    \bottomrule
  \end{tabular}
  }
  \caption{Zero-shot visual stimulus classification on the GOD dataset. The test set contains 50 categories that have no overlapping with the training set (top-1 chance=2.0\%). The results for CADA-VAE, MVAE, MMVAE, MoPoE-VAE, and BraVL are taken from~\cite{pami}. MindEye(211M) is a smaller version consisting of MLP backbone and projectors, with a residual block size of $1024\times1024$ (See Appendix B.3 for retrieval performance on the GOD dataset).}
  \vspace{-8mm}
  \label{tab2: zero-shot}
\end{table*}

As shown in Table \ref{tab2: main result}, compared to MLP Backbone, our DFT Backbone improves retrieval accuracy by 5\% and 14.9\% for two retrieval ways, indicating that the fine-grained representation of the image comes from the rich representation of the last hidden layer of CLIP rather than MLP's excessive attention to each voxel value. Note that our DFT Backbone is closest to MindEye's MLP Backbone with only contrastive learning, while Prior in MLP Backbone+Prior includes MSE loss for image reconstruction. 
\vspace{-0.5cm}
\subsection{LAION-5B retrieval}
Although the retrieval scale on NSD is large enough, we can still expand it to larger datasets, such as LAION-5B~\cite{schuhmann2022laion}. The final layer CLIP ViT-L/14 CLS embeddings for all 5 billion images are available at \url{https://knn.laion.ai/} and can be queried for K-nearest neighbor lookup via the CLIP Retrieval client~\cite{clipretrieval}. For LAION-5B retrieval, we train another similar \textbf{Lite-Mind-8M}, aligning voxels to the CLS embeddings of CLIP. For each test sample, we conduct a retrieval strategy in the same way as MindEye (first retrieve 16 candidate images using CLS embeddings, and the best image is selected based on having the highest cosine similarity to the fMRI voxel embeddings aligned to the final hidden layer of CLIP). Unlike MindEye, in the LAION-5B retrieval process, we used the Diffusion Projector to conduct image-to-image retrieval.

To compare the performance of large-scale dataset retrieval, we use metrics of image reconstruction for evaluation with previous image reconstruction methods. The experimental results are shown in Table \ref{tab2: LAION-5B}. The specific evaluation metrics in the table are as follows: PixCorr represents pixel-wise correlation between ground truth and retrieval/reconstruction images; SSIM is structural similarity index metric~\cite{ssim}; EfficientNetB1("Eff")~\cite{eff} and SwAV-ResNet50("SwAV")~\cite{swav} refer to average correlation distance; all other metrics refer to two-way identification (chance = 50\%).

Experiments show that our DFT Backbone performs better than MindEye in high-level metrics, as shown in Table \ref{tab2: LAION-5B}, although the backbone only has 8M parameters (MindEye-CLS backbone still has 125M. Note that both MindEye and Lite-Mind perform LAION-5B retrieval based on CLS alignment, resulting in a smaller number of parameters compared to full models). The results indicate that an efficient DFT backbone can also assist downstream retrieval tasks. The retrieval visualization results of Lite-Mind on LAION-5B retrieval for Subject 1 are shown in Figure \ref{fig:LAION-5B}, and the specific retrieval performance and visualization results of each Subject are shown in Appendix B.2.

\subsection{GOD zero-shot classification}
Since fMRI data are much fewer and shorter than those of the NSD dataset usually, we conducted zero-shot classification tasks on a smaller fMRI dataset to verify Lite-Mind's generalization ability. The voxel lengths collected by Subjects on the GOD dataset range from 4133 to 4643, approximately 30\% of NSD, and there are only 1200 fMRI-image pairs on the training dataset (4.8\% of NSD). 

We conduct zero-shot classification by using fMRI-to-image retrieval to find the specific image and using simple prompt text templates \textit{"{An image of [class]}"} to obtain the category of retrieved images(CLIP has a classification accuracy of 76.2\% on ImageNet without pre-training on it, so it can be seen as zero-shot classification). We train our \textbf{Lite-Mind-15.5M} to align voxels to images on the GOD dataset. We also replicated the MLP backbone and projector of MindEye, changing the residual block of $4096\times4096$ to $1024\times1024$ to accommodate the reduction of voxel length(We empirically found that the original $4096\times4096$ performed even worse), and MindEye still has 210M parameters. As shown in Table \ref{tab2: zero-shot}, the experimental results indicate that our proposed Lite-Mind still guarantees robustness and establishes a new state-of-the-art for zero-shot classification on the GOD dataset without complex text prompt templates or auxiliary text representation training. On the contrary, due to the sudden decrease in dataset size and voxel length, the performance of MindEye significantly decreases, indicating excessive reliance on training data size for voxel value-wise mappings with large MLP Backbone. More retrieval and classification results are provided in Appendix B.3.
\vspace{-2mm}

\subsection{Ablations and visualization}
We assess the effectiveness of different modules within Lite-Mind to investigate the lightweight and efficient performance in this session. All experiments below are for Subject 1 on the NSD dataset.
 
\textbf{Architectural Analysis.} We trained multiple DFT Backbone models with different depths to evaluate the impact of the number of layers of fMRI Spectrum Compression on retrieval efficiency(Table \ref{tab2: ablation result}). As the Filter Block layers gradually deepen, the number of model parameters increases, the image retrieval accuracy gradually improves, and the improvement speed tends to be gradual. We interestingly found that the accuracy of brain retrieval tends to converge faster than the image retrieval and only 6 layers of Filter Blocks are needed to approach the highest value. We have also conducted more ablation experiments on hyper-parameters to demonstrate the robustness of Lite-Mind. Refer to Appendix B.4 for it.

\begin{table}[h]
\vspace{-1mm}
  \centering
  \scalebox{0.8}{
  \begin{tabular}{{@{}ccccc@{}}}
    \toprule
    Depth& FLOPs(G)&Parameters & Image Retrieval & Brain Retrieval\\

    \midrule
     MindEye&5.66&996M&88.8\%&84.9\%\\
     1&0.05&0.7M&83.5\%&94.0\%\\
     6&0.22&3.4M&90.6\%&97.2\%\\
     12&0.43&6.6M&91.0\%&97.2\%\\
     21&0.80&12.5M&94.6\%&97.4\%\\
    \bottomrule
  \end{tabular}
  }
  \caption{Ablation experiments on fMRI Spectrum Compression revealed a positive correlation between the depth of the Filter Blocks and enhanced forward retrieval accuracy.}
  \vspace{-5mm}
  \label{tab2: ablation result}
\end{table}

\textbf{Module Analysis.} We conducted ablation experiments on two main modules of DFT Backbone in Table \ref{tab: w/o}. We completely removed the Filter blocks and aligned tokens directly to the image embeddings with frequency projector, achieving a forward retrieval accuracy of 65.8\%, indicating the importance of frequency domain filtering for denoising and compression. Then we replaced FreMLP with the FC layer in the real domain, where tokens are flattened into a tensor for alignment, similar to the last layer of the MLP backbone of MindEye. The model performance rapidly declined, indicating that tokens processed by DFT Backbone have better mapping performance with FreMLP in the frequency domain.

\begin{table}[h]
  \centering
  \vspace{-1mm}
  \scalebox{1.0}{
  \begin{tabular}{{@{}ccccc@{}}}
    \toprule
    Module&Image Retrieval & Brain Retrieval\\

    \midrule
     DFT Backbone&94.6\%&97.4\%\\
     w/o Filter Blocks&65.8\%&74.7\%\\
     w/o FreMLP&43.4\%&47.8\%\\
    \bottomrule
  \end{tabular}
  }
  \caption{Ablation experiments on modules of DFT Backbone.}
  \vspace{-6mm}
  \label{tab: w/o}
\end{table}




\textbf{Cerebral cortex.} To explore the impact of different cortical regions on retrieval accuracy, we used Takagi's~\cite{takagi2023high} method to extract the \textit{stream} ROI on the NSD dataset, covering the visual cortex of \textit{nsdgenal}, and trained the individual retrieval DFT Backbones for visual cortices (\textit{early}, \textit{lateral}, \textit{parietal}, \textit{ventral}). As shown in Table \ref{addtional brain region}, it can be observed that the early visual cortex has the greatest impact on retrieval accuracy. Although with a voxel length of 5917, Lite-Mind can still achieve retrieval accuracy of 85.0\% and 93.4\%, which is consistent with Takagi's research. They demonstrated that visual stimuli are almost dominated by the \textit{early} visual cortex and indicate that \textit{nsdgenal} still has a low signal-to-noise ratio, proving that fully connected MLP Backbones are unnecessary. Interestingly, for other cortical regions, both \textit{lateral} and \textit{ventral} have some impacts on retrieval accuracy, while \textit{parietal} shows little.

\begin{figure}[!t]
  \centering
   \includegraphics[width=1.0\linewidth]{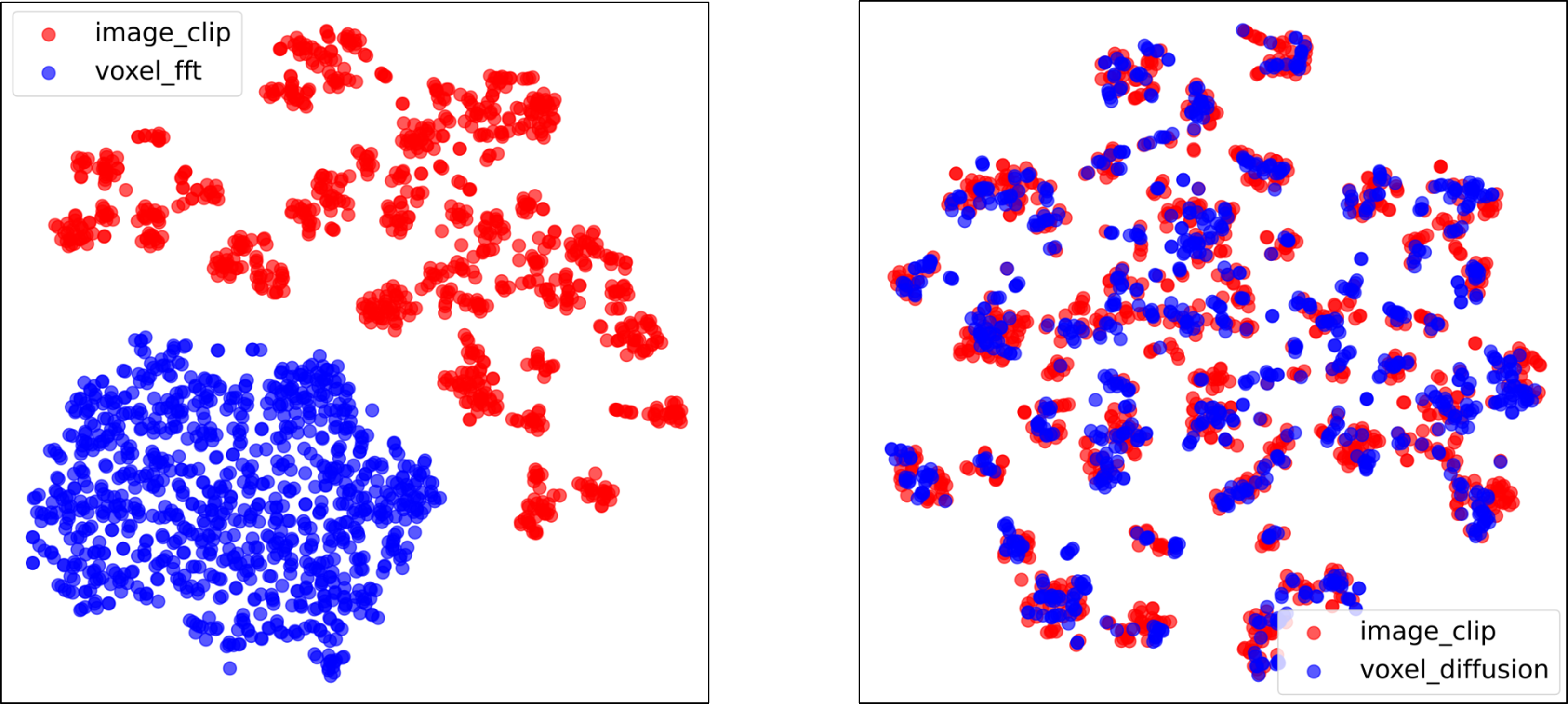}

   \caption{T-SNE of embeddings of 982 test fMRI by diffusion projector on LAION-5B retrieval for Subject 1. In the figure, image-clip represents image-clip CLS embeddings, while voxel-fft and voxel-diffusion represent voxel CLS embeddings by DFT Backbone or Diffusion Projector. The diffusion projector plays a role in bringing vectors closer, enabling image-to-image retrieval on the LAION-5B dataset.}
   \vspace{-1mm}
   \label{fig:prior}
\end{figure}

\textbf{Visualization.} We visualize the CLS embeddings for LAION-5B using T-SNE in Figure \ref{fig:prior}. The figure shows that the diffusion projector successfully transformed the voxel CLS embeddings learned from contrastive learning into the image CLS embeddings, and the two kinds of embeddings are well fused for image-to-image retrieval on the LAION-5B dataset. We have also visualized the alignment process of voxel embeddings by contrastive learning on the NSD dataset in Appendix C.3 and learned filters of Filter Blocks in fMRI Spectrum Compression in Appendix C.5.

\begin{table}[!t]
  \centering
  \vspace{-2mm}
  \scalebox{0.9}{
  \begin{tabular}{@{}ccccc@{}}
    \toprule
    \multirow{2}{*}{Region} & \multirow{2}{*}{Voxel Length}& \multirow{2}{*}{Parameters} & \multicolumn{2}{c}{Retrieval}\\
    \cline{4-5}
    & & & Image$\uparrow$&Brain$\uparrow$\\

    \midrule
     early&5917&12.4M&85.0\% & 93.4\%\\
     lateral&7799&12.4M&26.9\% & 29.0\%\\
     parietal&3548&12.3M&9.6\% & 11.2\%\\
     ventral&7604&12.3M&18.1\% & 22.4\%\\
    \bottomrule
  \end{tabular}
  }
  \caption{Retrieval performance with different cerebral cortex for Subject 1 on the NSD dataset.}
  \vspace{-6mm}
  \label{addtional brain region}
\end{table}

\section{Limitations}
Due to inherent differences between Subjects, Lite-Mind still needs to train Subject-specific models. Although our exploratory experiments have shown that a common model can achieve comparable retrieval accuracy on four Subjects, it's still necessary to ensure sufficient training data. 
\section{Conclusion}
In this work, we propose Lite-Mind, an extremely lightweight brain representation learning paradigm based on Fourier Transform for fMRI-to-image retrieval. Our DFT Backbone is an efficient means of obtaining fine-grained representation alignment between fMRI signals and visual stimuli. With high retrieval accuracy, Lite-Mind can transform downstream fMRI tasks into retrieval-based tasks, such as zero-shot classification. Meanwhile, Lite-Mind's dependence on dataset size and voxel length is less than that of larger models, demonstrating excellent robustness across different fMRI datasets.

\section{Acknowledgments}

This work is supported by the National Key Research and Development Program of China (No. 2022YFB3104700), the National Natural Science Foundation of China (No. 61976158, No. 62376198), Shanghai Baiyulan Pujiang Project (No. 08002360429). 


\bibliographystyle{ACM-Reference-Format}
\bibliography{sample-base}

\appendix
\setcounter{figure}{5}
\setcounter{table}{6}
\section{Additional Method Details}
\label{method details}
\subsection{Implementation details}
\begin{figure}[!t]
  \centering
  \begin{subfigure}{0.4\textwidth}
    \includegraphics[width=1.0\linewidth]{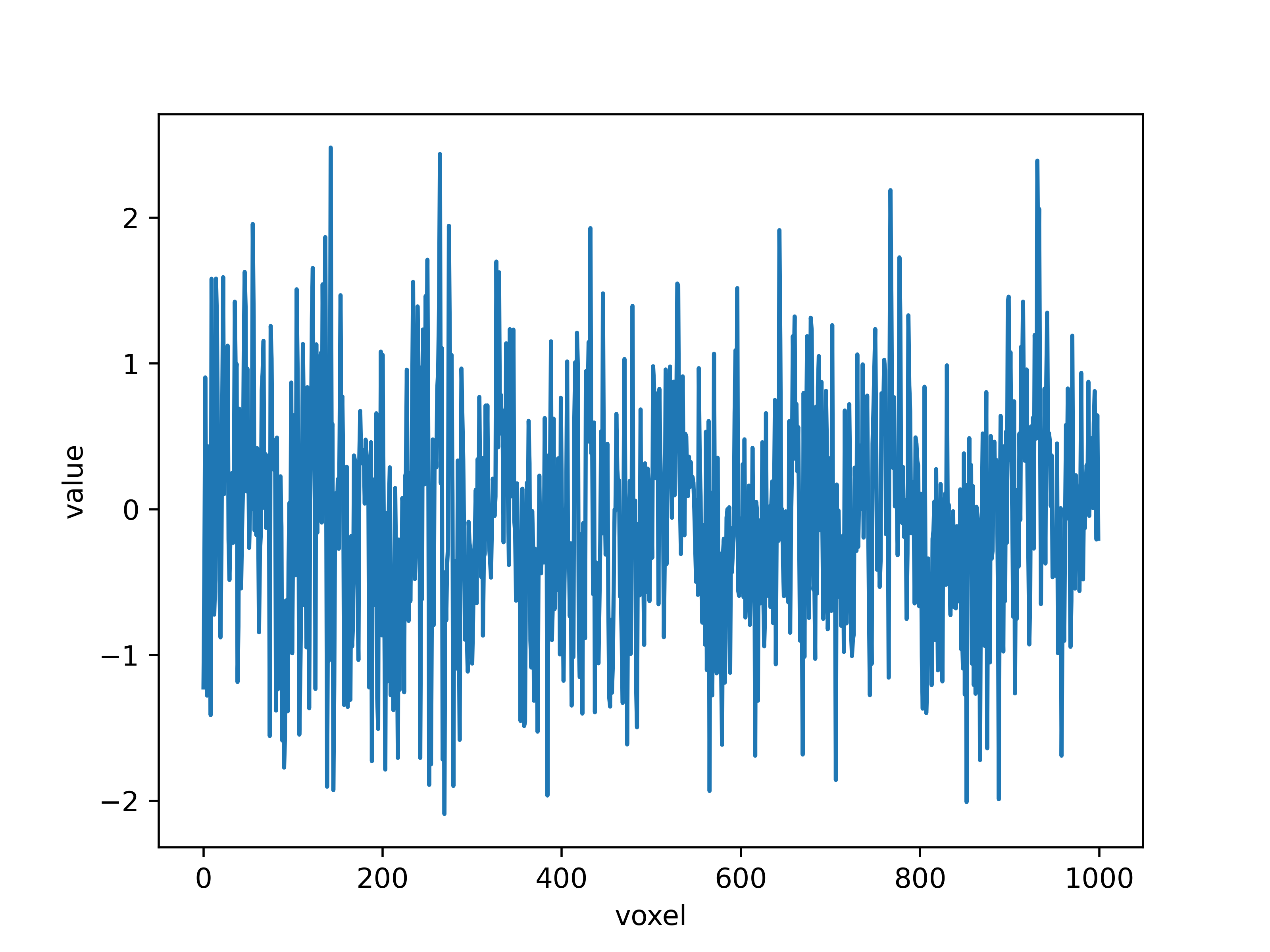}
    \label{fig:voxel_mindeye}
    \vspace{-3mm}
    \caption{MindEye}
  \end{subfigure}
  \begin{subfigure}{0.4\textwidth}
    \includegraphics[width=1.0\linewidth]{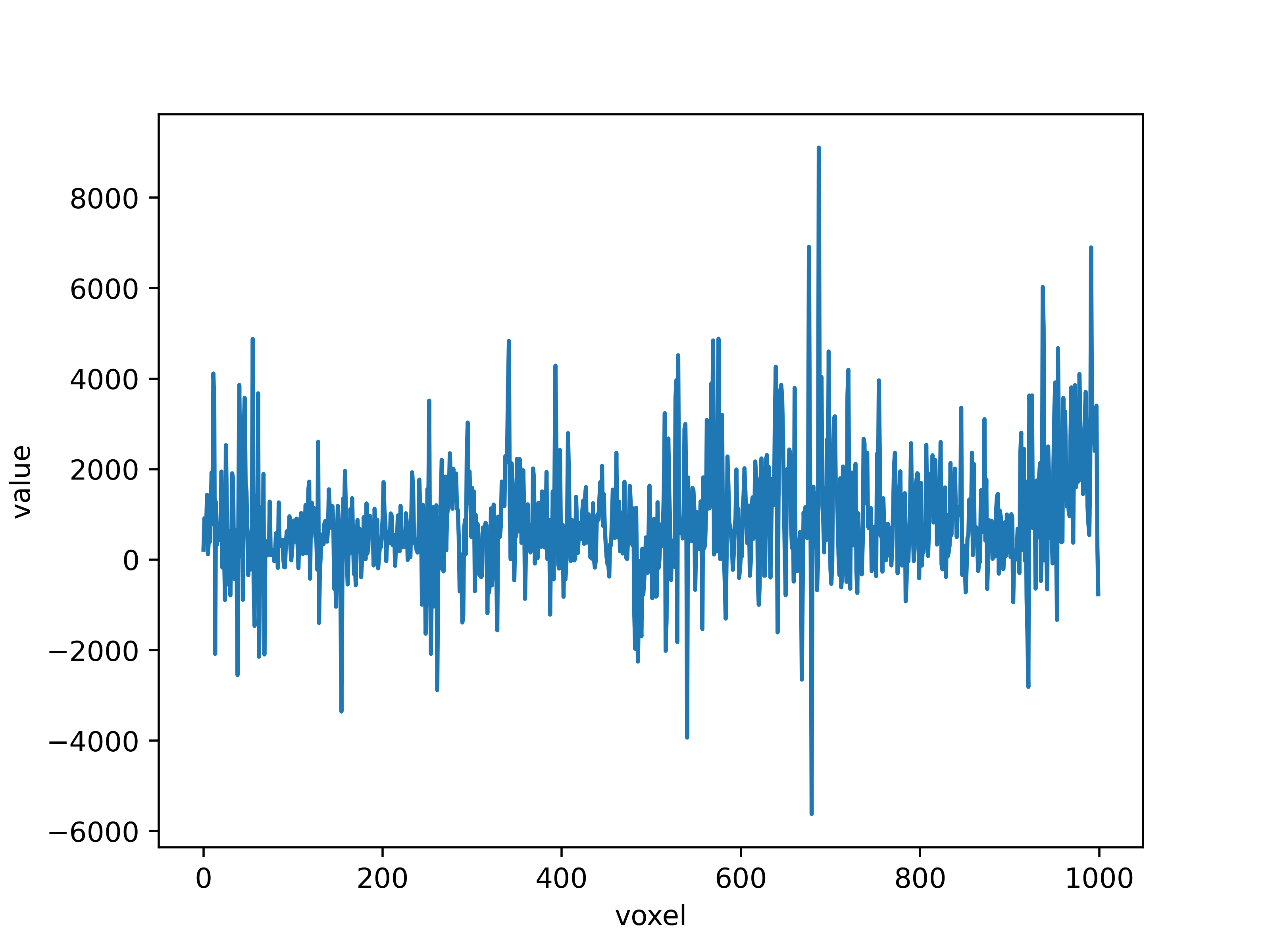}
    \label{fig:voxel_litemind}
    \vspace{-3mm}
    \caption{Lite-Mind}
    \vspace{-3mm}
  \end{subfigure}
  \caption{The fMRI averaged activities of MindEye and Lite-Mind responding to the same image, respectively. The figures only visualize the activities of the first 1000 voxels for illustration.}
  \label{fig:voxel}
\end{figure}

\textbf{Data preprocessing.} We downloaded the NSD dataset from the official website and used Takagi's code to extract \textit{nsdgenal} fMRI voxels, while Takagi extracted the \textit{stream} region of the NSD dataset. We noticed that MindEye scaled the fMRI voxel values in advance, while we did not. The difference in fMRI voxels input data is shown in the example in Figure \ref{fig:voxel}. NSD image files come from nsd-stimuli.hdf5 file and have a unified size of $425\times425$. We did not perform any data augmentation on the image and straightly extracted the hidden layer representation (size of $257\times768$) of the image through CLIP ViT-L/14 for training.

\textbf{Hyper-parameters.} On the NSD dataset, during training DFT Backbone, the weight decay is set to 7, $\tau$ is $1/e^8$, $\alpha=1$, and CLIP's contrastive loss is unidirectional for image retrieval and fMRI retrieval. Owning to the lightweight nature of Lite-Mind, the batch size is set to 500, the learning rate is 1e-3, patch size is set to 480, and Filter library size is 4. Filter block layers are the same for all subjects.

About LAION-5B retrieval, for DFT Backbone, the weight decay is set to 7, $\tau$ is $1/e^8$, while the weight decay is 6.02e-2 for diffusion projector. CLIP's contrastive loss is unidirectional for image retrieval, $\alpha=0.5$, the batch size is set to 80 and the learning rate is 1.16e-3 for DFT Backbone while 1.1e-4 for diffusion projector. Note that hyper-parameters in Diffusion Projector are the value of the open-source DALLE·2 from \url{https://github.com/lucidrains/DALLE2-pytorch}.

On the GOD dataset, the hyperparameters are the same as those on the NSD dataset, except for the batch size and patch size which are set to 1200 and 8 for all 5 subjects on the GOD dataset, respectively.

\section{Additional Experiment Results}
\subsection{Additional Results of Retrieval}
\label{addtional nsd results}
To demonstrate the applicability of our method, we also conducted experiments on three other Subjects, i.e., Subject 2, Subject 5, and Subject 7, on the NSD dataset, and the experimental results are shown in Table~\ref{tab4: addtional retrieval}. In order to control model size similarity, we did not make more targeted model adjustments for subjects with shorter voxel lengths. It can be seen that the retrieval accuracies of Subjects 1 and 2 are greater than 90\%, and the retrieval accuracies of Subjects 5 and 7 are also above 80\%. The results prove that DFT Backbone can efficiently work on different subjects. Note that MindEye only includes the results of the overall model in Subjects 2, 5, and 7 in its Appendix, and it does not evaluate the effect of individual MLP Backbone. As a result, Table~\ref{tab4: addtional retrieval} does not present any evaluation results for MindEye.

\begin{table}[h]
  \centering
  \scalebox{0.8}{
  \begin{tabular}{@{}ccccc@{}}
    \toprule
    \multirow{2}{*}{Method} & \multirow{2}{*}{Voxel Length}& \multirow{2}{*}{Parameters} & \multicolumn{2}{c}{Retrieval}\\
    \cline{4-5}
    & & & Image$\uparrow$&Brain$\uparrow$\\

    \midrule
     Lite-Mind(Subj 1)&15724&12.51M&94.6\% & 97.4\%\\
     Lite-Mind(Subj 2)&14278&12.49M&94.1\% & 98.2\%\\
     Lite-Mind(Subj 5)&13039&12.47M&80.5\% & 86.3\%\\
     Lite-Mind(Subj 7)&12682&12.47M&81.7\% & 82.3\%\\
    \bottomrule
  \end{tabular}
  }
  \caption{Addtional retrieval performance for individual subjects on 982 test images of the NSD dataset.}
  \label{tab4: addtional retrieval}
\end{table}

\begin{table*}[!t]
  \centering
  \begin{tabular}{@{}ccccccccc@{}}
    \toprule
    \multirow{2}{*}{Method} & \multicolumn{4}{c}{Low-Level}& \multicolumn{4}{c}{High-Level}\\
    \cmidrule(lr){2-5} \cmidrule(lr){6-9}
     &PixCorr$\uparrow$&SSIM$\uparrow$&Alex(2)$\uparrow$&Alex(5)$\uparrow$&Incep$\uparrow$&CLIP$\uparrow$&Eff$\downarrow$&SwAV$\downarrow$\\

    \midrule
     Lite-Mind(Subject 1)&.134&.332&78.8\%&88.9\%&88.5\%&88.8\%&.730&.451\\
     Lite-Mind(Subject 2)&.120&.328&78.0\%&89.4\%&86.3\%&87.4\%&.730&.446\\
     Lite-Mind(Subject 5)&.123&.332&79.4\%&90.0\%&88.8\%&89.9\%&.712&.440\\
     Lite-Mind(Subject 7)&.121&.331&78.7\%&88.8\%&87.8\%&88.5\%&.723&.448\\
     \bottomrule
  \end{tabular}
  \caption{LAION-5B retrieval alternative reconstruction performance for the specific subject.}
  \label{tab5: laion5b subjects}
\end{table*}
\subsection{Additional Results of LAION-5B Retrieval}
\label{addtional laion5b results}
\begin{figure*}[p]
  \centering
  \begin{subfigure}{1.0\textwidth}
    \includegraphics[width=\linewidth]{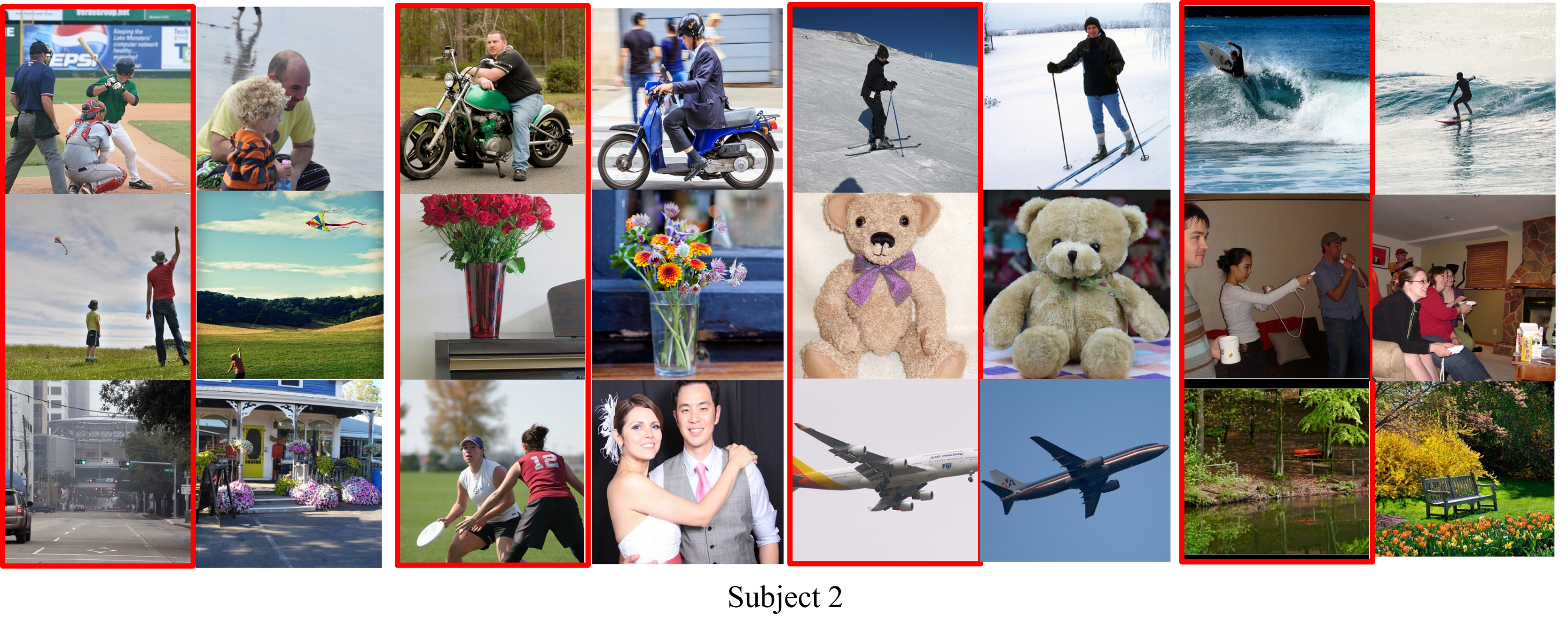}
    \label{fig:subj02}
  \end{subfigure}
  \begin{subfigure}{1.0\textwidth}
    \includegraphics[width=\linewidth]{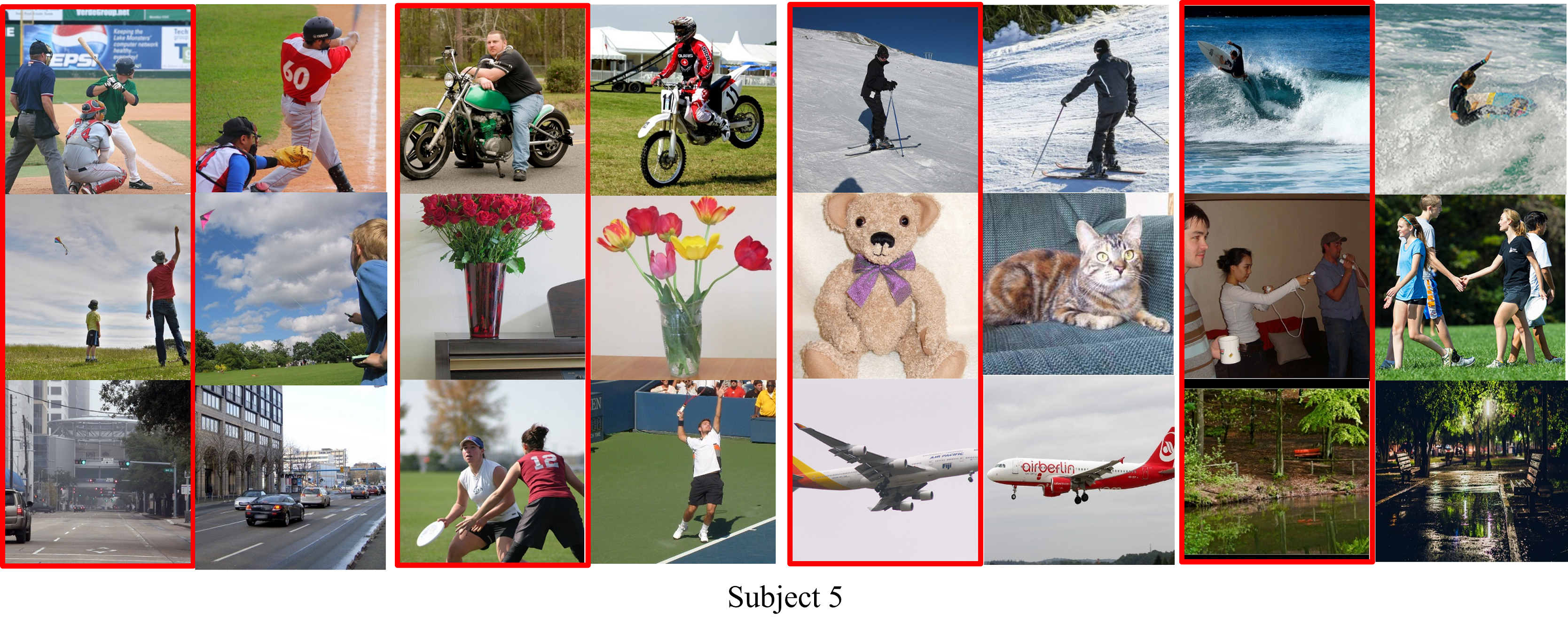}
    \label{fig:subj05}
  \end{subfigure}
  \begin{subfigure}{1.0\textwidth}
    \includegraphics[width=\linewidth]{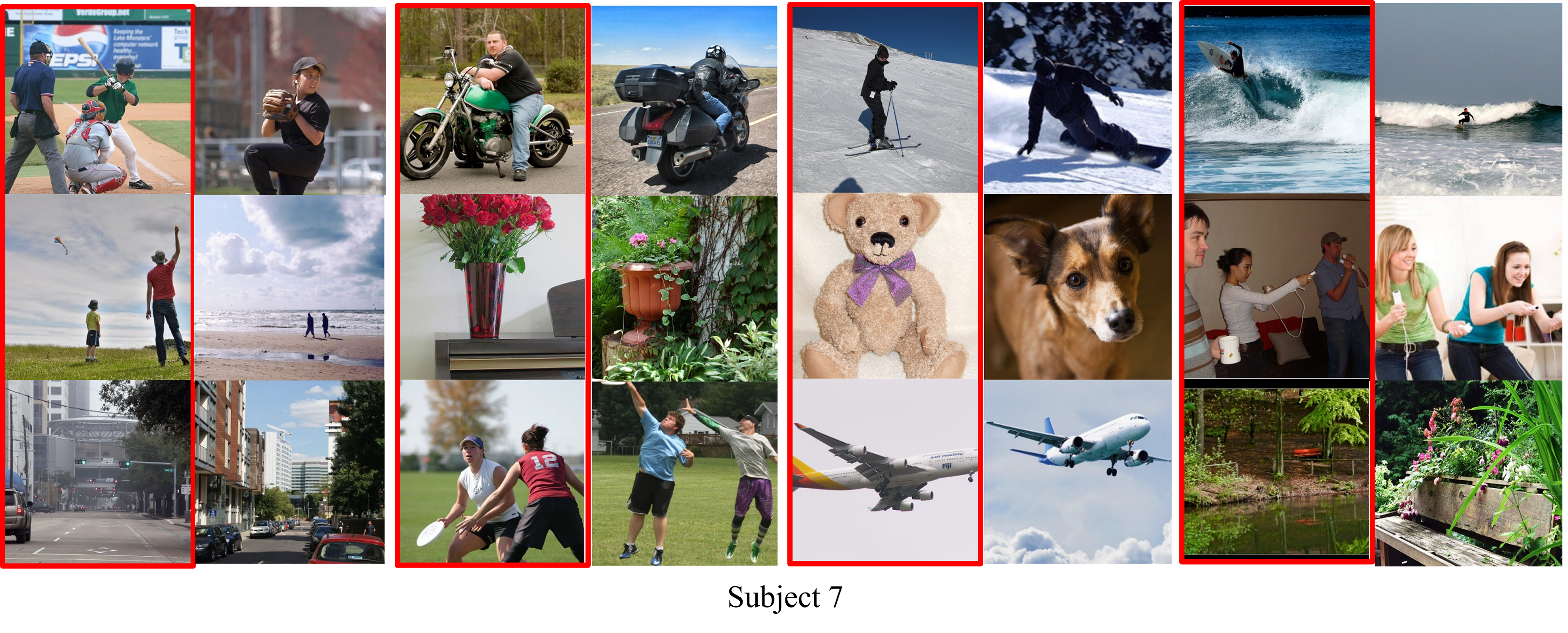}
    \label{fig:subj07}
  \end{subfigure}
  \vspace{-1cm}
  \caption{Addtional retrieval results corresponding to Figure 4. The left column marked by a red box in every two columns represents the original image seen by the subject, and the right column represents the image retrieved on LAION-5B.}
  \label{fig:addtional laion5b}
\end{figure*}

In order to better reflect the retrieval performance of Lite-Mind on LAION-5B, we presented the performance indicators of LAION-5B retrieval substitution reconstruction for other subjects, i.e., Subject 2, Subject 5, and Subject 7, in Table \ref{tab5: laion5b subjects}, corresponding to the average performance of subjects in Table 1. Similarly, the visualization results of the other subjects in Figure \ref{fig:addtional laion5b} correspond to the image samples of Subject 1 in Figure 4. Based on the comprehensive table and graph, it can be found that Lite-Mind has good generalization on all four subjects, verifying the LAION-5B retrieval ability of Lite-Mind on different subjects. Meanwhile, as shown in Figure \ref{fig:addtional laion5b}, the retrieval performance of LAION-5B completely depends on the retrieval accuracy of the CLS model in the test set, such that images retrieved incorrectly in the test set may also have retrieval bias on LAION-5B, for example treating a teddy bear as an image of a cat or dog as shown in Figure \ref{fig:addtional laion5b}. However, both MindEye and Lite-Mind exhibit relatively low retrieval accuracy with aligned CLS embedding models. In the future, it would be beneficial to explore models that improve the alignment of CLS embeddings or employ more efficient methods to directly perform retrieval through hidden layers in LAION-5B.

\subsection{Additional Results of Zero-shot Classification}
\label{addtional god results}
In the main body of the paper, we only demonstrated the zero-shot classification effect of Lite-Mind on the GOD dataset. The corresponding retrieval results of each Subject are shown in the Table \ref{tab6: addtional god} below.
\begin{table}[h]
  \centering
  \scalebox{0.8}{
  \begin{tabular}{@{}ccccc@{}}
    \toprule
    \multirow{2}{*}{Method} & \multirow{2}{*}{Voxel Length}& \multirow{2}{*}{Parameters} & \multicolumn{2}{c}{Image Retrieval$\uparrow$}\\
    \cline{4-5}
    & & &top1&top5\\

    \midrule
     Lite-Mind(Subj 1)&4466&15.50M&30.0\% & 60.0\%\\
     Lite-Mind(Subj 2)&4404&15.46M&38.0\% & 58.0\%\\
     Lite-Mind(Subj 3)&4643&15.64M&38.0\% & 72.0\%\\
     Lite-Mind(Subj 4)&4133&15.24M&42.0\% & 62.0\%\\
     Lite-Mind(Subj 5)&4370&15.43M&26.0\% & 54.0\%\\
    \bottomrule
  \end{tabular}
  }
  \caption{Retrieval performance on the GOD dataset.}
  \label{tab6: addtional god}
\end{table}

\subsection{Hyper-parameters Experiment}
\label{hyper-parameters}
In this section, we explore the influence of some hyperparameters on the model, including patch size and number of filters $M$, to verify the parameter sensitivity of the model. All the experimental results on the NSD dataset are from Subject 1, with a retrieving pool size of 300.

\textit{Patch size.} We conducted experiments by varying the patch size as presented in Table \ref{patch}. The results exhibit stable retrieval accuracy above 90\%, which indicates that our DFT Backbone is not sensitive to part size. Since we found patch size has less effect on retrieval accuracy, we chose a relatively large patch size (i.e., 480) to ensure fewer parameters and faster convergence.
\begin{table}[h]
  \centering
  \scalebox{1.0}{
  \begin{tabular}{@{}ccccc@{}}
    \toprule
    \multirow{2}{*}{Patch size} & \multirow{2}{*}{Parameters} & \multicolumn{2}{c}{Retrieval}\\
    \cline{3-4}
    & & Image$\uparrow$&Brain$\uparrow$\\

    \midrule
    50&14.0M&91.6\%&96.4\%\\
    200&12.7M&92.9\%&96.9\%\\
    480&12.5M&94.6\%&97.1\%\\
    600&12.3M&94.2\%&97.3\%\\
    900&11.9M&92.1\%&95.5\%\\
    \bottomrule
  \end{tabular}
  }
  \caption{Retrieval performance for different patch size on the NSD dataset.}
  \label{patch}
\end{table}

\textit{Filter library size.} We conducted experiments by varying the Filter library size as presented in Table \ref{filters}. Specifically, there is a significant performance improvement from $M = 1$ to $M = 4$, while a slight fluctuation is observed for $M=4/8$. By setting $M = 4$, the model has the ability to acquire diverse and distinct feature patterns from various dimensions of the frequency response while still maintaining an appropriate computational cost. Therefore, we determine that $M = 4$ is the optimal choice on the NSD dataset.

\begin{table}[h]
  \centering
  \scalebox{1.0}{
  \begin{tabular}{@{}ccccc@{}}
    \toprule
    \multirow{2}{*}{Filters} & \multirow{2}{*}{Parameters} & \multicolumn{2}{c}{Retrieval}\\
    \cline{3-4}
    & & Image$\uparrow$&Brain$\uparrow$\\

    \midrule
    1&9.0M&89.8\%&96.4\%\\
    2&10.2M&93.5\%&97.4\%\\
    4&12.5M&94.6\%&97.1\%\\
    8&17.2M&94.4\%&96.7\%\\
    \bottomrule
  \end{tabular}
  }
  \caption{Retrieval performance for different Filter library size on the NSD dataset.}
  \label{filters}
\end{table}

\textbf{Embedding Dimension.}We conducted additional experiments to verify the impact of embedding dimensions on retrieval accuracy, as shown in Table \ref{tab: representation size}. Among them, two different CLIP models (i.e., CLIP ViT/B-32 and CLIP ViT/L-14) were used to extract image embeddings, with a total of four embedding dimensions, by CLS embeddings and the last hidden layer, respectively. From the results, it can be seen that the retrieval accuracy of Lite-Mind is higher when the embedding dimensions of the image are longer and the representation of the image is richer. The results verify that the fine-grained alignment we mentioned above comes from the rich representation of the image, rather than the fMRI voxel value fully connected heavy MLP Backbone.

\begin{table}[ht]
  \centering
  \vspace{-1mm}
  \scalebox{1.0}{
  \begin{tabular}{@{}ccccc@{}}
    \toprule
    \multirow{2}{*}{Embeddings} & \multirow{2}{*}{Dimension} & \multirow{2}{*}{Parameters} & \multicolumn{2}{c}{Retrieval}\\
    \cline{4-5}
    && & Image$\uparrow$&Brain$\uparrow$\\

    \midrule
     ViT-B/32 CLS&512&6.7M&57.5\% & 61.8\%\\
     ViT-L/14 CLS&768&6.7M&60.3\% & 64.2\%\\
     ViT-B/32 Hidden&50×512&10.4M&91.1\% & 96.5\%\\
     ViT-L/14 Hidden&257×768&12.5M&94.6\% & 97.4\%\\
    \bottomrule
  \end{tabular}
  }
  \caption{Retrieval performance with different CLIP embeddings for Subject 1 on the NSD dataset.}
  \vspace{-6mm}
  \label{tab: representation size}
\end{table}

\begin{figure*}[!t]
  \centering
   \includegraphics[width=1.0\linewidth]{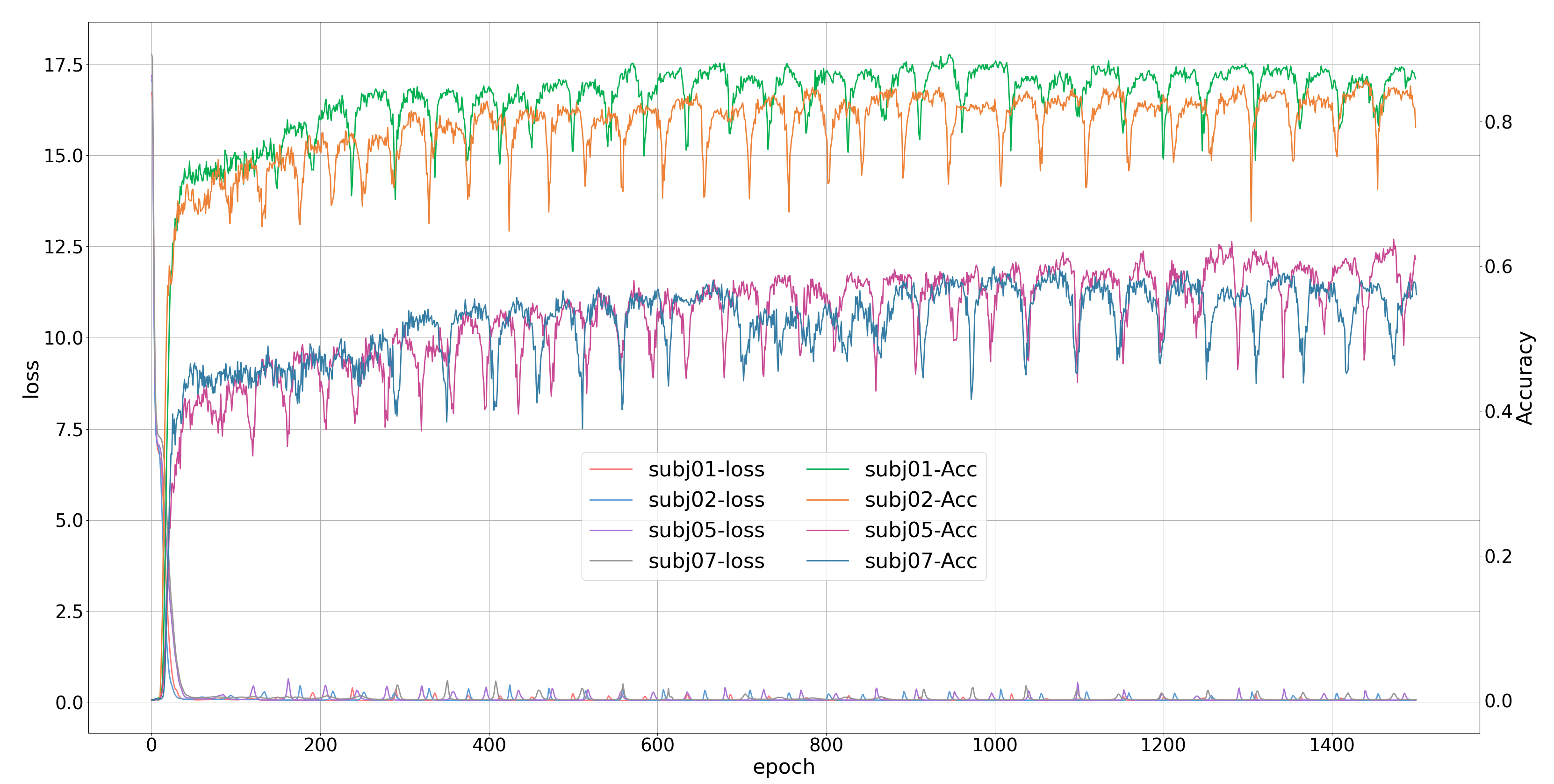}
   \caption{Lite-Mind's training loss curve and testing accuracy curve for Suject 1, 2, 5, 7 on the NSD dataset. The testing accuracy is calculated based on a retrieval pool of size 982.}
   \label{fig:nsd_loss}
\end{figure*}

\begin{figure}[!t]
  \centering
   \includegraphics[width=1.0\linewidth]{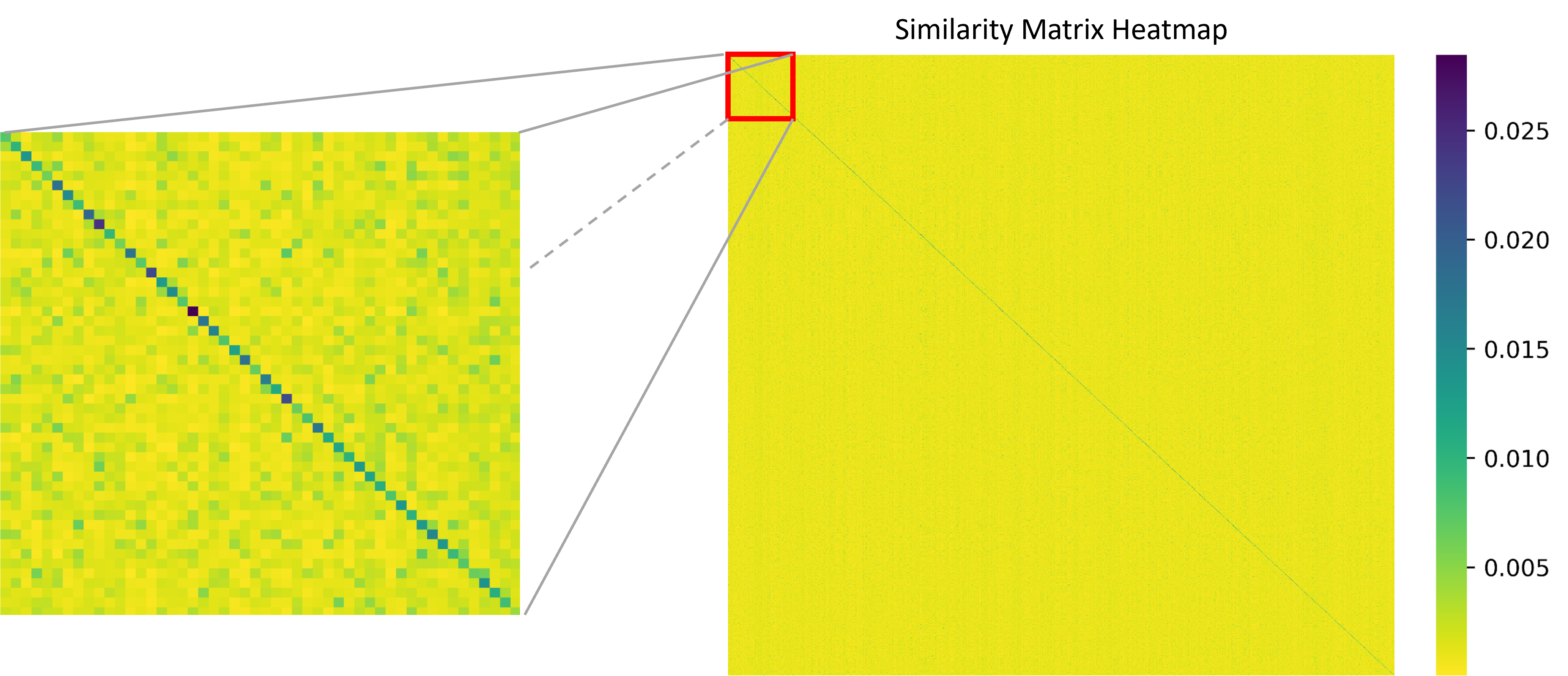}
   \caption{Lite-Mind's retrieval heatmap on the NSD dataset for Subject 1. The larger figure on the right represents 982 × 982's heatmap, and the smaller figure on the left represents the 50 × 50 subgraph.}
   \label{fig:retrieval heatmap}
\end{figure}

\section{Additional Visualization}
\label{additional visualization}
\subsection{Training Curve}
We visualized the training process of all four subjects on the NSD dataset in Figure \ref{fig:nsd_loss}. As the training epochs increased, the loss of the training set rapidly decreased, while the accuracy of the test set rapidly increased and then showed a slow upward trend. Accuracy refers to the hit rate of correct retrieval from 982 test set images.

\subsection{Retrieval Heatmap}
We visualized the retrieval heatmap for Subject 1 on all 982 test images of the NSD dataset in Figure \ref{fig:retrieval heatmap}. It can be observed that the similarity is highest on the diagonal, and the color of the retrieved heat map is darker. It shows that Lite-Mind has effectively retrieved corresponding images, even if there are many similar images in the test set, which verifies the fine-grained ability of Lite-Mind.

\subsection{Information Alignment}
\label{information}
We visualized the T-SNE plot between the voxel embeddings output by DFT Backbone and the image embeddings of frozen CLIP as the accuracy of the test set improved, as shown in Figure \ref{fig:information2}. We can observe that as the training progresses, the retrieval accuracy of the test set improves, and the shape of voxel embeddings tends to be closer to image embeddings, indicating the success of contrastive learning.
\subsection{More Retrieval Cases}
\label{more cases}
\begin{figure*}[!t]
  \centering
  \vspace{-3mm}
   \includegraphics[width=1.0\linewidth]{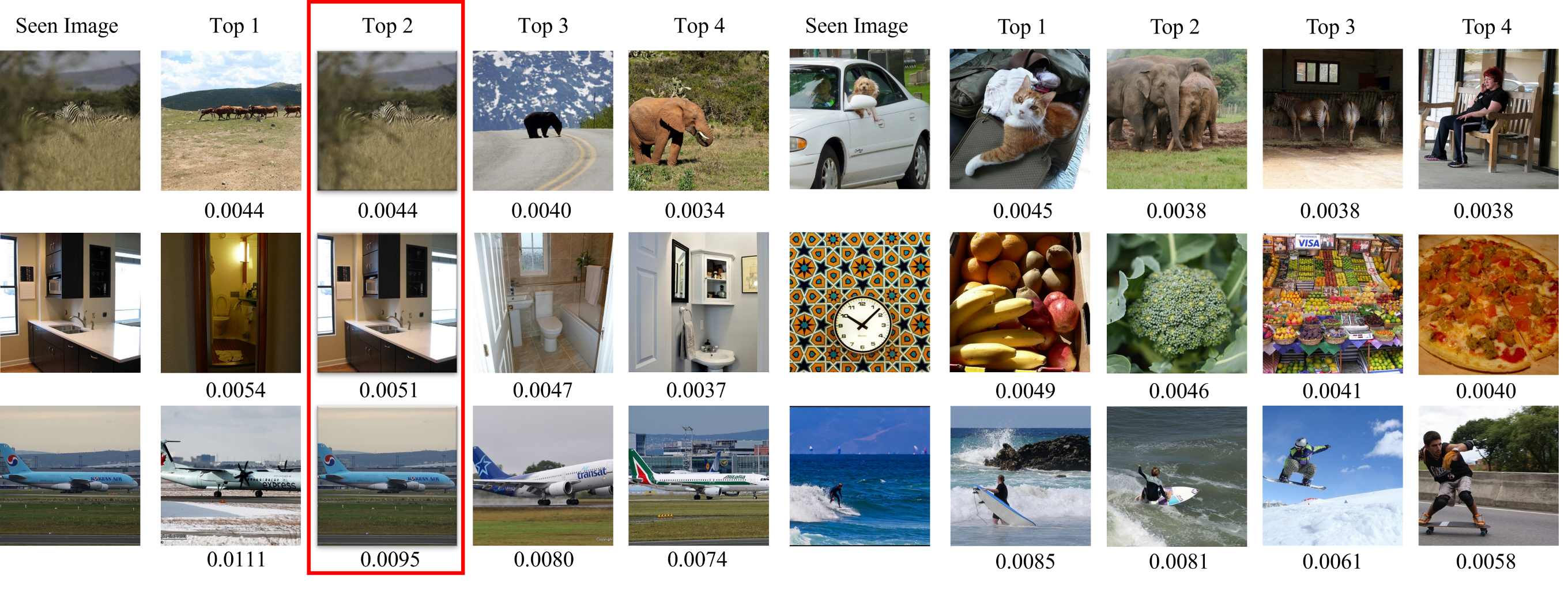}
\vspace{-6mm}
   \caption{Partial failure retrieval results of Lite-Mind on all 982 test images for Subject 1. The number below each image represents the similarity score.}
   \vspace{-4mm}
   \label{failure}
\end{figure*}

In this section, we visualize retrieval failure cases in Figure \ref{failure}, although only dozens of images are not in the Top 1. From the left half of the Figure, it can be seen that though not in Top 1, Lite-Mind still retrieved ground-truth in Top 2, and images in Top 4 are similar(either semantically similar or structurally similar, eg. animals in the wild or an airplane on the runway). On the contrary, a smaller proportion of ground-truth did not appear within the Top 4, as shown in the right half of the Figure. From the perspective of the images themselves, most of the scenes are too complex, even abstract (as shown in the second image), which may be the reason for the retrieval failure.

\subsection{Visualization in Frequency Domain}
\label{visualization filters}
\begin{figure}
  \centering
  \begin{subfigure}{0.4\textwidth}
    \includegraphics[width=1.0\linewidth]{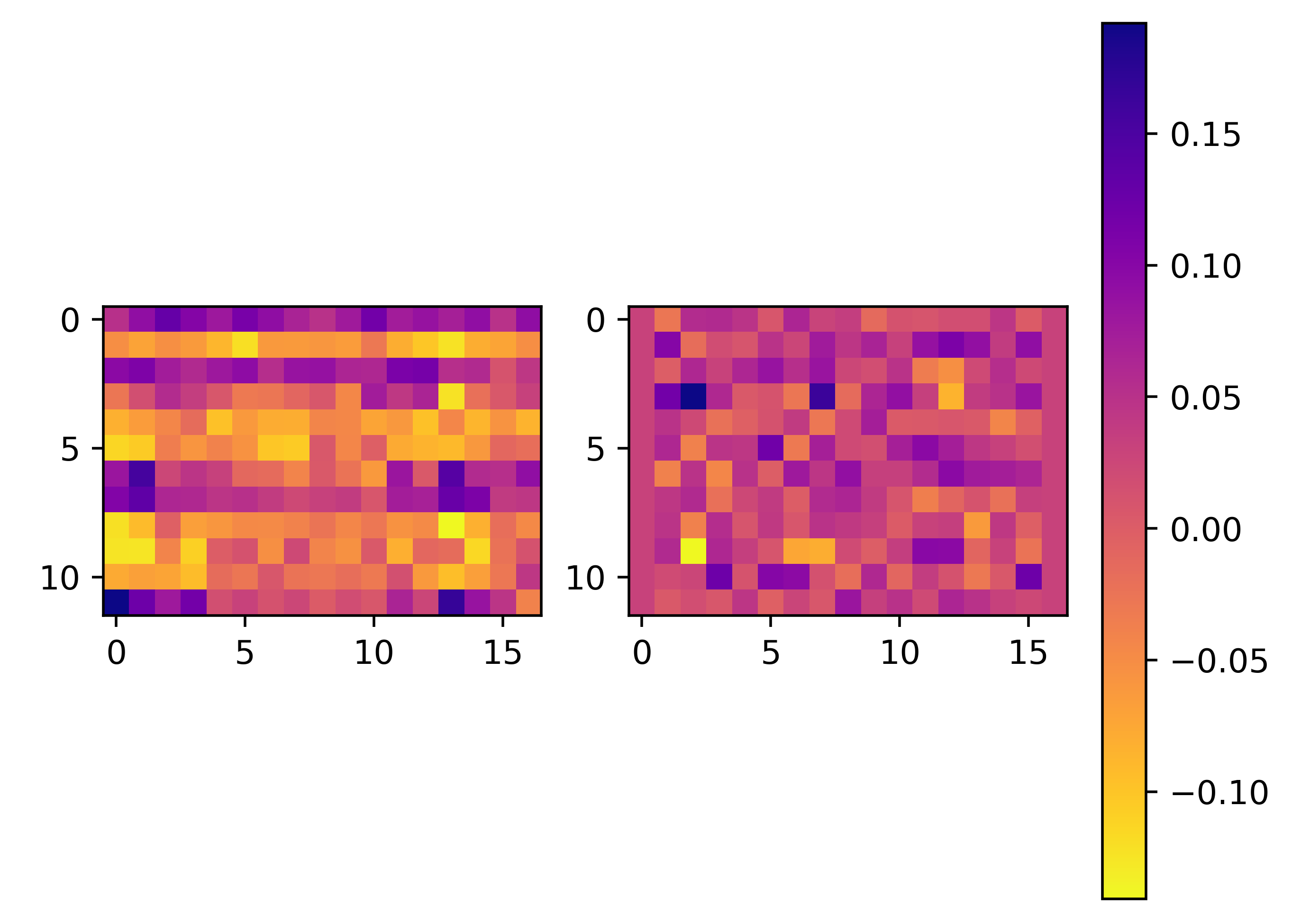}
    \label{Filter for CLS token channel.}
    \caption{The real part and imaginary of a filter weights.}
  \end{subfigure}
  \begin{subfigure}{0.4\textwidth}
    \includegraphics[width=1.0\linewidth]{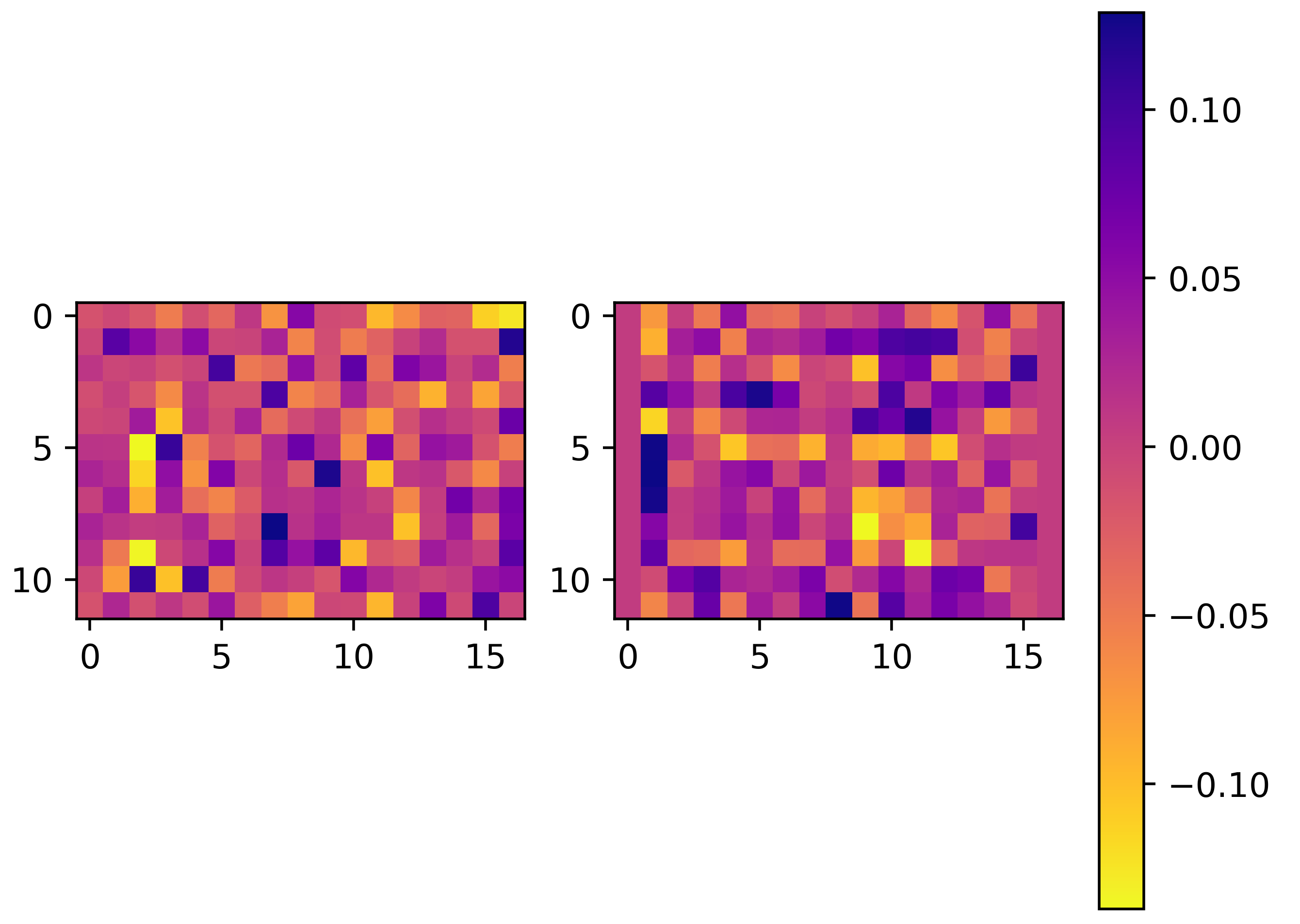}
    \label{fig:filter-140}
    \caption{The real part and imaginary of another filter weights from the same Filter Library.}
  \end{subfigure}
  \caption{Weights visualization for DFT Backbone of $257\times768$ embedding length.}
  \label{fig:filter_hidden}
\end{figure}

\begin{figure}
  \centering
  \begin{subfigure}{0.4\textwidth}
    \includegraphics[width=1.0\linewidth]{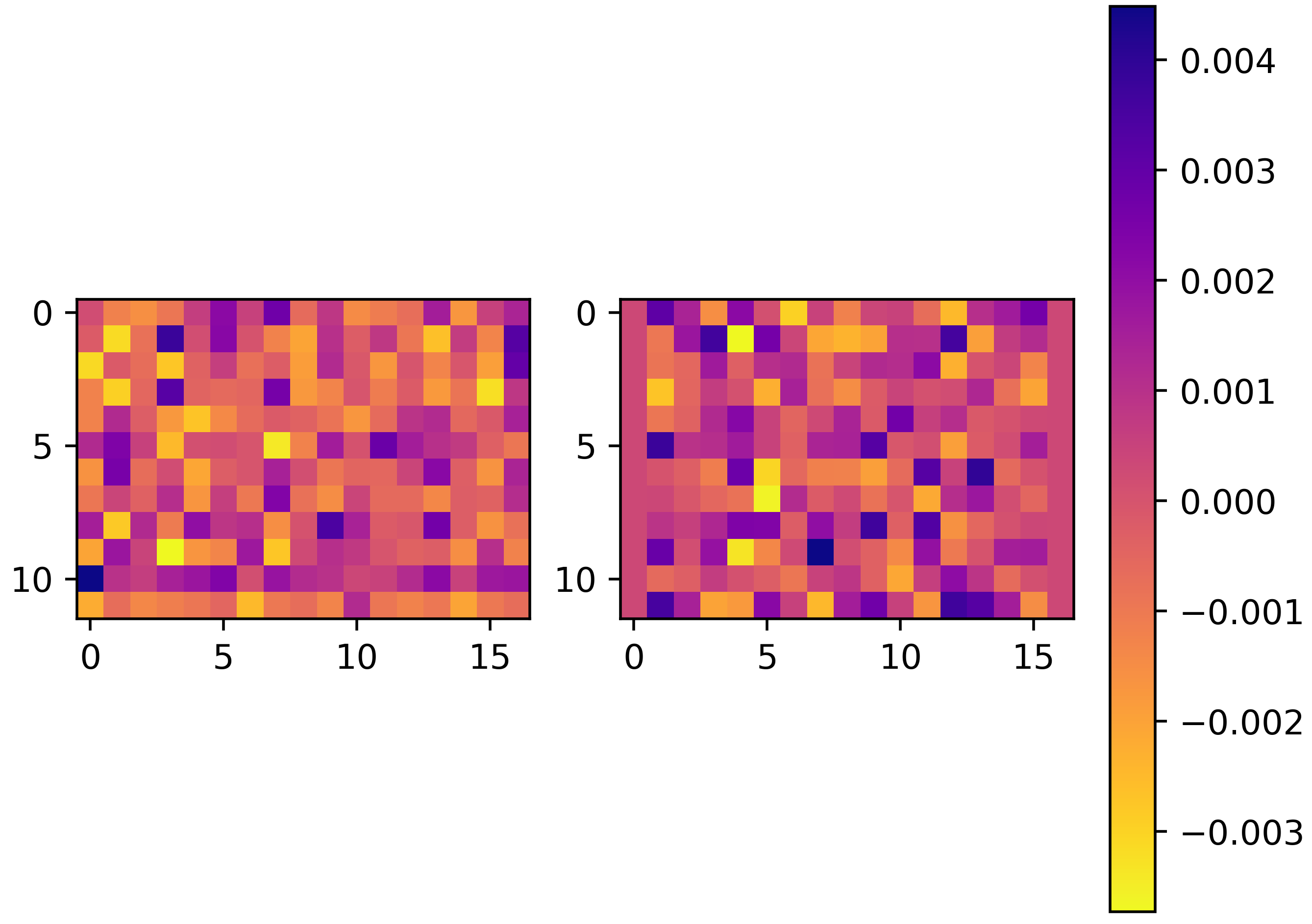}
    \label{Filter for CLS token channel.}
    \caption{The real part and imaginary of a filter weights.}
  \end{subfigure}
  \begin{subfigure}{0.4\textwidth}
    \includegraphics[width=1.0\linewidth]{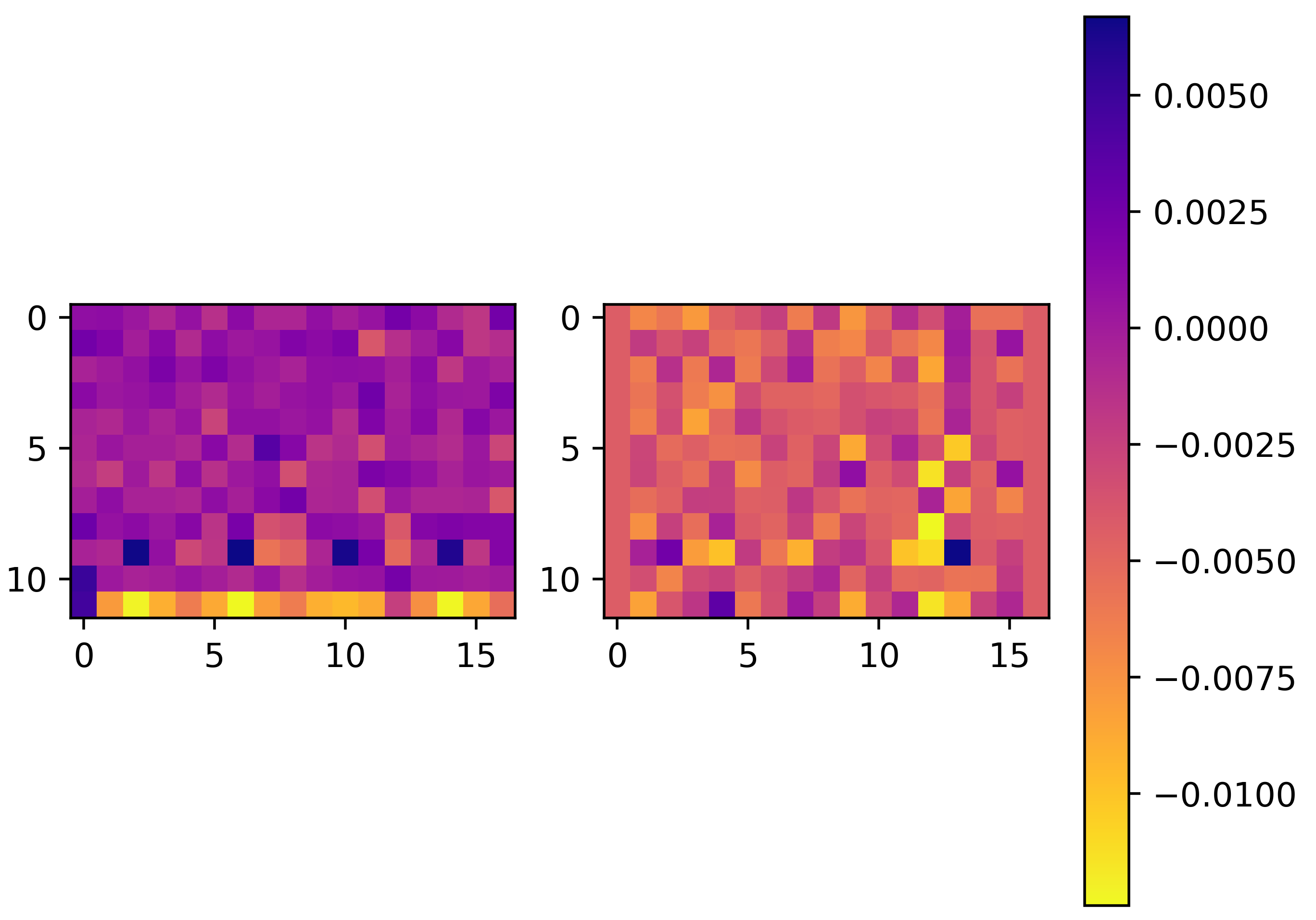}
    \label{fig:filter-140}
    \caption{The real part and imaginary of another filter weights from the same Filter Library.}
  \end{subfigure}
  \caption{Weights visualization for DFT Backbone of 768 CLS embedding length.}
  \label{fig:filter_CLS}
\end{figure}

\begin{figure*}[!t]
  \centering
   \includegraphics[width=1.0\linewidth]{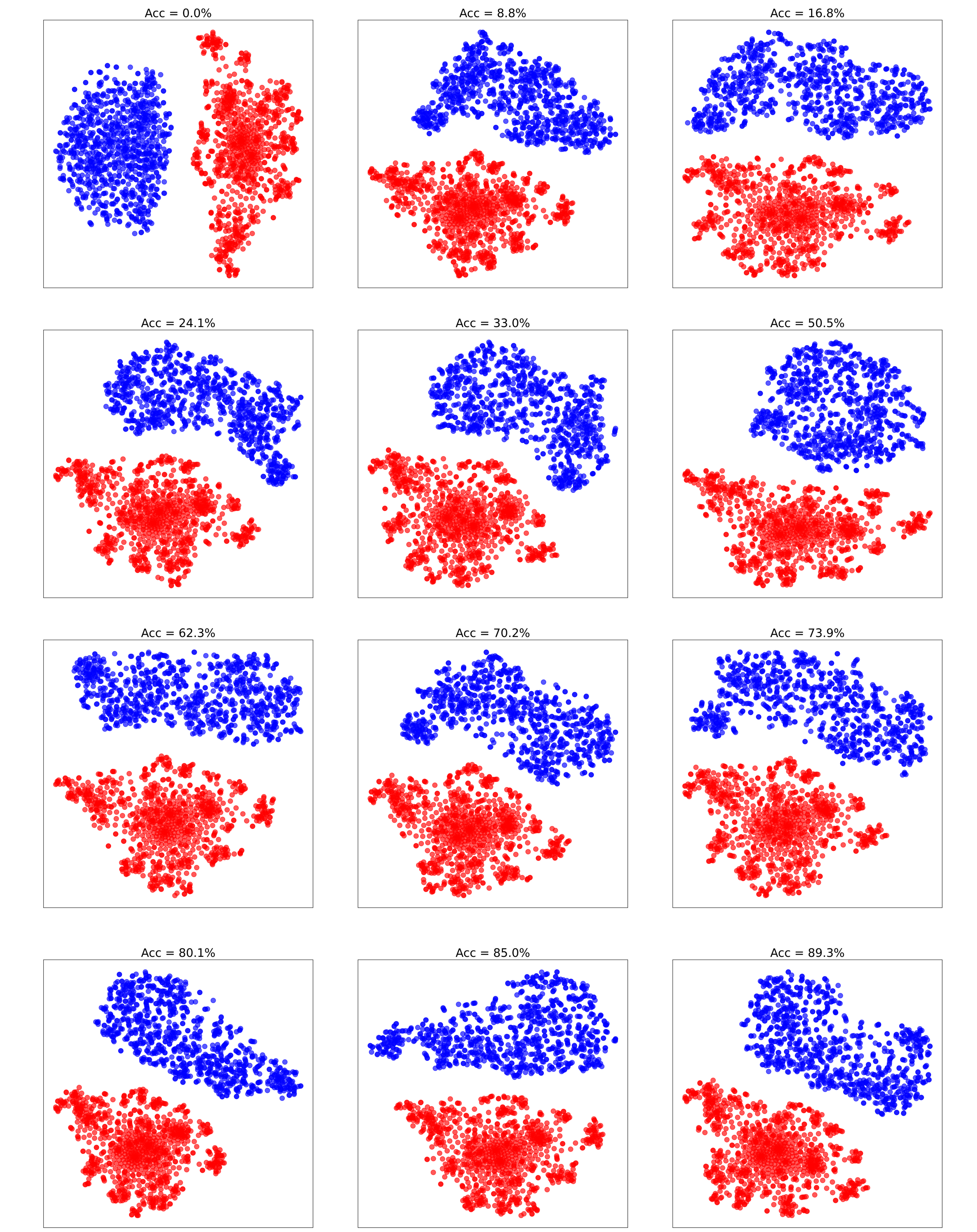}
   \caption{T-SNE visualization between the voxel embeddings output by DFT Backbone and the image embedding of frozen CLIP. Accuracy in the title refers to the hit rate of correct retrieval from 982 test set images and the blue dots represent voxel embeddings.}
   \label{fig:information2}
\end{figure*}

We visualized the weights of Filter library of Filter Blocks, as shown in Figure \ref{fig:filter_hidden} and \ref{fig:filter_CLS}. Visualization is divided into the real part (left) and imaginary part (right) of filter weights. It can be observed that different filters have varying degrees of attention to different tokens, and the frequency domain better captures this characteristic. Interestingly, the weight of the imaginary part for the first and last tokens is almost always 0, indicating that the noise is distributed in these two tokens.

\section{Theoretical Analysis.}
\subsection{Complex Multiplication.}
\label{complex}
For two complex number values $\mathcal{Z}_1=(a+jb)$ and $\mathcal{Z}_2=(c+jd)$, where $a$ and $c$ is the real part of $\mathcal{Z}_1$ and $\mathcal{Z}_2$ respectively, $b$ and $d$ is the imaginary part of $\mathcal{Z}_1$ and $\mathcal{Z}_2$ respectively. Then the multiplication of $\mathcal{Z}_1$ and $\mathcal{Z}_2$ is calculated by:
\begin{equation}
    \begin{aligned}
    \mathcal{Z}_1\mathcal{Z}_2&=(a+jb)(c+jd)\\
    &=ac+j^2bd+jad +jbc\\
    &=(ac-bd)+j(ad+bc)
    \end{aligned}
\end{equation}
\subsection{Theorem Proof.}
\label{theorm}
\textbf{Theorem 1.} Suppose that $\textbf{H}$ is the representation of raw fMRI voxel tokens and $\mathcal{H}$ is the corresponding frequency components of the spectrum, then the energy of voxel tokens in the spatial domain is equal to the energy of its representation in the frequency domain. Formally, we can express this with the above notations:
\begin{equation}
  \int_{-\infty}^{\infty} {|\textbf{H}(v)|}^2\,\mathrm{d}v = \int_{-\infty}^{\infty} {|\mathcal{H}(f)|}^2\,\mathrm{d}f
\end{equation}
Where $\mathcal{H}(f) = \int_{-\infty}^{\infty}{|\textbf{H}(v)|}e^{-j2{\pi}fv}\,\mathrm{d}v$, $v$ is the token dimension, $f$ is the frequency dimension.

\begin{proof}
Given the representation of raw voxel token series $H \in \mathbb{R}^{C \times N}$, let us consider performing integration in the $N$ dimension (spatial dimension), denoted as the integral over $v$, then
\begin{equation}
  \int_{-\infty}^{\infty} {|\textbf{H}(v)|}^2\,\mathrm{d}v = \int_{-\infty}^{\infty} {\textbf{H}(v)}{\textbf{H}^*(v)}\,\mathrm{d}v
\end{equation}
where $\textbf{H}^*(v)$ is the conjugate of $\textbf{H}(v)$. According to IDFT, $\textbf{H}^*(v) = \int_{-\infty}^{\infty}{\mathcal{H}^*(f)}e^{-j2{\pi}fv}\,\mathrm{d}f$, we can obtain

\begin{equation}  
  \begin{aligned}
  \int_{-\infty}^{\infty} {|\textbf{H}(v)|}^2\,\mathrm{d}v &= \int_{-\infty}^{\infty} {\textbf{H}(v)}[\int_{-\infty}^{\infty}{\mathcal{H}^*(f)}e^{-j2{\pi}fv}\,\mathrm{d}f]\,\mathrm{d}v\\
  &= \int_{-\infty}^{\infty} {\mathcal{H}}^*(f)[\int_{-\infty}^{\infty}{|\textbf{H}(v)|}e^{-j2{\pi}fv}\,\mathrm{d}v]\,\mathrm{d}f\\
  &= \int_{-\infty}^{\infty} {\mathcal{H}^*(f)}{\mathcal{H}(f)}\,\mathrm{d}f\\
  &= \int_{-\infty}^{\infty} {|\mathcal{H}(f)|}^2\,\mathrm{d}f
  \end{aligned}
\end{equation}
Proved.
\end{proof}

Therefore, the energy of a voxel token series in the spatial domain is equal to the energy of its representation in the frequency domain.

\textbf{Theorem 2.} Given the voxel token series input $\textbf{H}$ and its corresponding frequency domain conversion $\textbf{H}$, the operations of frequency-domain MLP on $\textbf{H}$ can be represented as global
convolutions on $\textbf{H}$ in the spatial domain. This can be given by:
\begin{equation}
    \mathcal{H}\mathcal{W}+\mathcal{B} = \mathcal{F}(\textbf{H}*W + B)
\end{equation}
Where $\mathcal{F}$ is DFT, $*$ is a circular convolution, $\mathcal{W}$ and $\mathcal{B}$ are the complex number weight and bias, and $W$ and $B$ are the weight and bias in the spatial domain.
\begin{proof}
Suppose that we conduct operations in the N (i.e., token dimension), then
\begin{equation}
    \mathcal{F}(\textbf{H}(v)*W(v))=\int_{-\infty}^{\infty}{(\textbf{H}(v)*W(v))e^{-j2\pi fv}}\,\mathrm{d}v
\end{equation}
According to convolution theorem, $\textbf{H}(v)*W(v)=\int_{-\infty}^{\infty}(\textbf{H}(\tau)W(v-\tau))\,\mathrm{d}\tau$, then
\begin{equation}
    \begin{aligned}
    \mathcal{F}(\textbf{H}(v)*W(v))&=\int_{-\infty}^{\infty}{\int_{-\infty}^{\infty}{(\textbf{H}(\tau)W(v-\tau))e^{-j2\pi fv}}\,\mathrm{d}\tau}\,\mathrm{d}v\\
    &=\int_{-\infty}^{\infty}{\int_{-\infty}^{\infty}{W(v-\tau)e^{-j2\pi fv}}\,\mathrm{d}v}\textbf{H}(\tau)\,\mathrm{d}\tau
    \end{aligned}
\end{equation}
Let $x=v-\tau$,then
\begin{equation}
    \begin{aligned}
    \mathcal{F}(\textbf{H}(v)*W(v))&=\int_{-\infty}^{\infty}{\int_{-\infty}^{\infty}{W(x)e^{-j2\pi f(x+\tau)}}\,\mathrm{d}x}\textbf{H}(\tau)\,\mathrm{d}\tau\\
    &=\int_{-\infty}^{\infty}{\int_{-\infty}^{\infty}{W(x)e^{-j2\pi fx}e^{-j2\pi f\tau}}\,\mathrm{d}x}\textbf{H}(\tau)\,\mathrm{d}\tau\\
    &=\int_{-\infty}^{\infty}{(\textbf{H}(\tau)*)e^{-j2\pi f\tau}}\,\mathrm{d}\tau \int_{-\infty}^{\infty}{(\textbf{W}(x)*)e^{-j2\pi fx}}\,\mathrm{d}x\\
    &=\mathcal{H}(f)\mathcal{W}(f)
    \end{aligned}
\end{equation}
Accordingly, $\textbf{H}(v)*W(v)$ in the spatial domain is equal to $\mathcal{H}(f)\mathcal{W}(f)$ in the frequency domain. Therefore, the operations of FreMLP ($\mathcal{H}\mathcal{W}+\mathcal{B}$) in the token dimension (i.e., v = N ) are equal to the operations ($\mathcal{F}(\textbf{H}*W + B)$) in the spatial domain. This implies that frequency-domain MLPs can be viewed as global convolutions in the spatial domain. Proved.
\end{proof}

\subsection{Complexity Analysis}
For a fMRI voxel with a length of $l$, we divide it into $n$ patches. Assuming $L_1$ and $L_2$ are the layer depths of MLP Backbone and DFT Backbone respectively, the middle layer dimension of MLP Backbone is $D$, and the alignment embedding dimension is $n'\times D'$, where $n'$ is the the number of tokens of CLIP. The time complexity of MLP Backbone is $O(lD+L_1D^2+n'DD')$. For DFT Backobone, the time complexity of patchify and tokenization is $O(lD')$, and the time complexity of DFT, IDFT, and filtering for each layer is $O(2nD'\operatorname{log} n+nD')$. The time complexity of FreMLP is $O(2nD'\operatorname{log} n+2nn'D'+2n'D')$. Thus the time complexity of the entire DFT Backbone is:
\begin{equation}
\begin{aligned}
    O(4nD'\operatorname{log} n+(n+2nn'+2n')D')\\
    =O((n\operatorname{log} n+nn'+n')D')\\
\end{aligned}
\end{equation}

Quantitative analysis algorithm complexity for DFT Backbone has been shown in Table 3.


\end{document}